\pdfoutput=1
\documentclass[11pt]{article}

\usepackage[]{acl}

\usepackage{times}
\usepackage{latexsym}

\usepackage[T1]{fontenc}
\usepackage[utf8]{inputenc}

\usepackage{microtype}

\usepackage{inconsolata}

\usepackage{graphicx}
\usepackage{amsmath}
\usepackage{amsfonts}
\usepackage{booktabs}
\usepackage{multirow}
\usepackage{subcaption}
\usepackage{pifont} % using \ding{number} to selct
\usepackage{algorithm}
\usepackage{algpseudocode}
\usepackage{tikz}
\usepackage{makecell}
\usepackage{xspace}

\definecolor{blue_color}{RGB}{219,255,255}
\definecolor{yellow_color}{RGB}{255,255,173}

\newcommand{\md}{sNeuron-TST\xspace}

\title{Style-Specific Neurons for Steering LLMs in Text Style Transfer}

\author{Wen Lai$^{1,2}$, Viktor Hangya$^{2,3}$, Alexander Fraser$^{1,2}$\\\\
	$^1$ School of Computation, Information and Technology, Technical University of Munich, Germany \\ 
        $^2$Munich Center for Machine Learning, Germany \\
        $^3$Center for Information and Language Processing, LMU Munich, Germany \\
        {\tt \{wen.lai, alexander.fraser\}@tum.de, hangyav@cis.lmu.de}
        }

\begin{document}
\maketitle
\begin{abstract}
Text style transfer (TST) aims to modify the style of a text without altering its original meaning.
Large language models (LLMs) demonstrate superior performance across multiple tasks, including TST.
However, in zero-shot setups, they tend to directly copy a significant portion of the input text to the output without effectively changing its style.
To enhance the stylistic variety and fluency of the text, we present \textit{\md}, a novel approach for steering LLMs using \underline{s}tyle-specific \underline{neurons} in \underline{TST}.
Specifically, we identify neurons associated with the source and target styles and deactivate source-style-only neurons to give target-style words a higher probability, aiming to enhance the stylistic diversity of the generated text.
However, we find that this deactivation negatively impacts the fluency of the generated text, which we address by proposing an improved contrastive decoding method that accounts for rapid token probability shifts across layers caused by deactivated source-style neurons.
Empirical experiments demonstrate the effectiveness of the proposed method on six benchmarks, encompassing formality, toxicity, politics, politeness, authorship, and sentiment\footnote{\url{https://github.com/wenlai-lavine/sNeuron-TST}}.
\end{abstract}

\section{Introduction}
\label{sec:intro}
Text style transfer (TST;~\citealp{jin-etal-2022-deep,hu2022text}) aims to transform text from a source style to a target style while maintaining the original content and ensuring the fluency of the generated text.
Given any text $x$ in an original style $s_1$, the objective of TST is to transform $x$ into a new text $\hat{x}$ in a different style $s_2$ ($s_2 \neq s_1$), ensuring that the content remains unchanged despite the shift in style.
Large language models (LLMs;~\citealp{minaee2024large}) exhibit exceptional performance across various NLP tasks~\cite{chang2024survey}, including TST~\cite{ostheimer2023text,chen2024lmstyle}.
However, existing LLMs (e.g., LLaMA-3~\citealp{llama3}) tend to prioritize preserving the original meaning over enhancing stylistic differences in TST.
Our analysis reveals that 34\% of the outputs generated by LLaMA-3 are identical to the input text when tasked with transferring polite text to impolite text (Section~\ref{sec:copy}).
Enhancing the generation of words that align with the target style during the decoding process remains a significant challenge in TST.

Recent LLMs have been successfully applied to TST, broadly categorized into two approaches:
(i) employing single-style or parallel-style text data for either full-parameter or parameter-efficient fine-tuning~\cite{mukherjee2024are,mukherjee2024text}, and 
(ii) leveraging the robust in-context learning capabilities of LLMs to create specialized prompts for zero-shot or few-shot learning~\cite{chen2024lmstyle,pan2024unsupervised}.
However, (i) typically requires substantial data and computational resources to achieve good results, while (ii) is highly sensitive to prompts, where even minor changes can significantly impact the outcomes~\cite{chen-etal-2023-relation}.

Neuron analysis~\cite{xiao2024exploring}, which aims to identify and understand the roles of individual neurons within a neural network, is a crucial method for enhancing the interpretability of neural networks and has garnered increasing attention in recent years.
By identifying neurons associated with specific attributes such as language~\cite{zhao2024large}, knowledge~\cite{niu2024does}, and skill~\cite{wang-etal-2022-finding-skill}, neuron analysis can boost performance on targeted tasks.
Recent research has demonstrated that focusing on language-specific neurons can markedly enhance the multilingual capabilities of LLMs during the decoding stage~\cite{kojima2024multilingual,tan2024neuron}.
However, the exploration of style-specific neurons remains relatively underexplored until now.

Thus motivated, we raise the following two research questions:\\
\textbf{Q1:} Do LLMs possess neurons that specialize in processing style-specific text? \\
\textbf{Q2:} If such neurons exist, how can we optimize their utilization during the decoding process to steer LLMs in generating text that faithfully adheres to the target style?

To address these research questions, we introduce \emph{\md}, a novel framework designed to steer LLMs in performing TST by leveraging style-specific neurons.
Initially, we feed both source- and target-style texts into the LLM to identify neurons that exclusively activate in each style based on their activation values.
We distinguish neurons active in both styles as overlapping neurons.
Notably, eliminating these overlapping neurons during style-specific neuron selection is crucial as their presence can hinder the generation of text in the target style.
Our experiments highlight that deactivating neurons specific solely to the source style (excluding those active in both source and target styles) improves style transfer accuracy while impacting sentence fluency.
Furthermore, to improve the fluency of generated text, we adapt the state-of-the-art contrastive decoding algorithm (Dola;~\citealp{chuang2024dola}) for optimal performance in TST tasks.
Our empirical findings (detailed in Section~\ref{sec:our_adapt}) reveal that layers primarily responsible for style-related outputs are concentrated in the model's latter layers, termed as \emph{style layers}.
This indicates that the determination of style-specific words predominantly occurs in these style layers.
More precisely, we refine the probability distribution of generated words by comparing logits from these style layers with the final layers, which exert significant influence on style-related outputs.

We conduct a comprehensive evaluation to verify the efficacy of our approach across six benchmarks: formality~\cite{rao-tetreault-2018-dear}, toxicity~\cite{logacheva-etal-2022-paradetox}, politics~\cite{voigt-etal-2018-rtgender}, politeness~\cite{madaan-etal-2020-politeness}, authorship~\cite{xu-etal-2012-paraphrasing} and sentiment~\cite{shen2017style}.
Each benchmark contains two distinct styles, resulting in a total of $12$ TST directions.
Experimental results demonstrate that our method generates a higher proportion of words in the target style compared to baseline systems, achieving superior style transfer accuracy and fluency, while preserving the original meaning of the text.

In summary, we make the following contributions:
\textbf{(i)}~To the best of our knowledge, this is the first work on using style-specific neurons 
to steer LLMs in performing text style transfer tasks.
\textbf{(ii)}~We emphasize the significance of eliminating overlap between neurons activated by source and target styles, a methodological innovation with potential applications beyond style transfer.
\textbf{(iii)}~We introduce an enhanced contrastive decoding method inspired by Dola. Our approach not only increases the production of words in the target style but also ensures the fluency of the generated sentences, addressing issues related to direct copying of input text in TST.
\section{Related Work}
\label{sec:related_work}

\noindent\textbf{Text Style Transfer.}
Recently, LLMs have shown promising results in TST through additional fine-tuning~\cite{mukherjee2024are,mukherjee2024multilingual,mukherjee2024text,dementieva-etal-2023-exploring}, in-context learning~\cite{chen2024lmstyle,zhang2024distilling,pan2024unsupervised,mai2023prefix} techniques or prompt-based text editing approaches~\cite{luo-etal-2023-prompt,liu2024adaptive}.
However, these methods often require either extensive computational resources or sensitive prompts, impacting their practicality.
In this paper, we focus on a novel decoding approach to guide LLMs for TST using fixed prompts and therefore it does not require significant computational consumption and ensures stable outputs.

\noindent\textbf{Neuron Analysis.}
Neuron analysis~\cite{xiao2024exploring} has emerged as a powerful method for elucidating the inner workings of neural network models, offering deeper insights into their behaviors and attracting growing interest in recent years.
The common practice is to associate neuron activation with learned knowledge, demonstrating effectiveness in tasks such as knowledge enhancement~\cite{li2024inference}, sentiment analysis~\cite{tigges2023linear} and multilingualism in LLMs~\cite{kojima2024multilingual,tan2024neuron}.
Motivated by the promising outcomes of neuron analysis in enhancing multilingual capabilities of LLMs, this paper posits the presence of style-specific neurons, identifies them, and integrates neuron activation and deactivation seamlessly into the decoding process.
%% whole framework
\begin{figure*}[thb]
\centering
	\includegraphics[width=\linewidth]{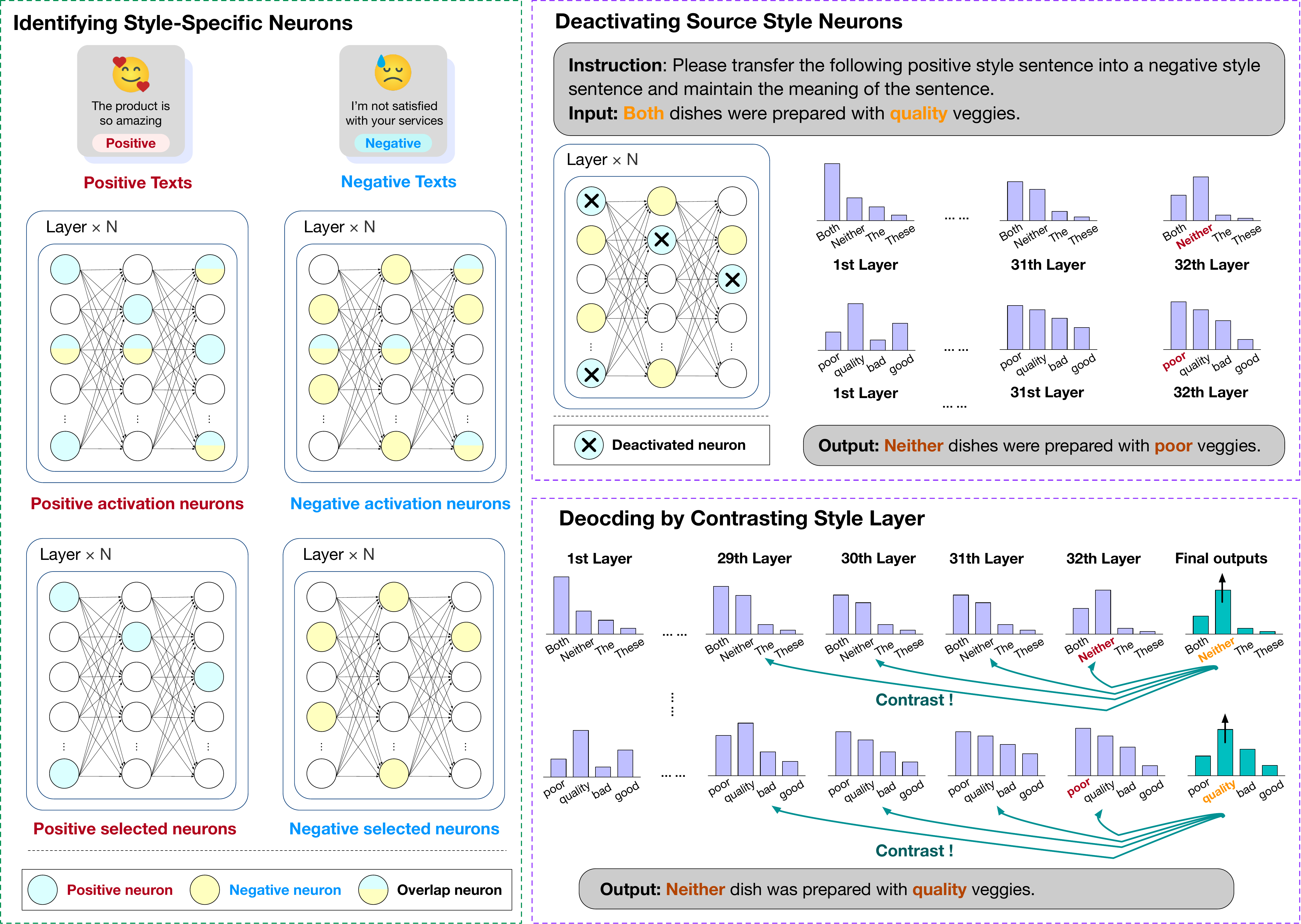}
	\caption{\label{fig:framework} Method overview. The whole framework consists of three parts: identifying style-specific neurons, deactivating source style neurons, and decoding by contrasting style layer. 
    The histogram represents the probability distribution of each word across different layers.
    When source style neurons are deactivated, LLMs tend to generate all target-style words, such as ``Neither'' and ``poor''. By employing contrastive decoding, LLMs take fluency into account and reduce the probability of generating ``poor''.
 }
 \vspace{-1.2em}
\end{figure*}

\section{Method}
\label{sec:method}
Our goal is to identify style-specific neurons to steer LLMs towards generating vocabulary tailored exclusively to a target style, while maintaining fluent text generation in a zero-shot setting.
To accomplish this, we first identify style-specific neurons based on their activation values and demonstrate the necessity of eliminating source- and target-style neurons to avoid overlap (Section~\ref{sec:identify}).
Then, we deactivate neurons associated solely with the source style, observing an increased probability of generating words aligned with the target style, albeit at the expense of fluency (Section~\ref{sec:deact}). Finally, we adapt the recent contrastive decoding approach Dola~\cite{chuang2024dola} to TST, ensuring the fluency of generated sentences (Section~\ref{sec:contras_decode}).
Figure~\ref{fig:framework} illustrates the framework of our approach.

\subsection{Identifying Style-Specific Neurons}
\label{sec:identify}
Neurons are commonly perceived as feature extractors that map neural networks to human-interpretable concepts~\cite{dreyer2024pure}.
However, neurons can exhibit polysemy, where a single neuron may encode multiple features (e.g., formal and informal styles), thereby complicating their interpretability.
To selectively modify specific features of LLMs without unintended changes, it becomes imperative to identify and remove unambiguous neurons.

\subsubsection{Neurons in LLMs}
The dominant architecture of LLMs is the Transformer~\cite{vaswani2017attention}, characterized by multiple layers of multi-head self-attention and feed-forward network (FFN) modules.
FFNs contain $2/3$ of the model's parameters and encode extensive information, which is crucial for multiple tasks~\cite{yang2024evolutionary}.
Moreover, the activation or deactivation of neurons within the FFN can exert significant influence on the model's output~\cite{garde2023deepdecipher}.
Inspired by this, we aim to identify neurons in the FFN modules of LLMs that are dedicated to specific styles.

Formally, the activation values of layer $j$ in a network are defined as:
\vspace{-0.2em}
\begin{equation}\label{eq:act_val}
    \vspace{-0.2em}
    \begin{aligned}
        a^{(j)} = \text{act\_fn}(W^{(j)} a^{(j-1)} + b^{(j)})
    \end{aligned}
\end{equation}
\noindent
where $W^{(j)}$ and $b^{(j)}$ are the weights and biases of layer $j$, while $a^{(j-1)}$ is the activation values of the previous layer and $\text{act\_fn}(\cdot)$ denotes the activation function (e.g., GLU;~\citealp{shazeer2020glu} used in LLaMA).
The $i^{th}$ neuron of the layer is considered to be active when its activation value $a_i^{(j)}>0$.

%% overlap statistics
\begin{figure}[thp]
\centering
	\includegraphics[width=\linewidth]{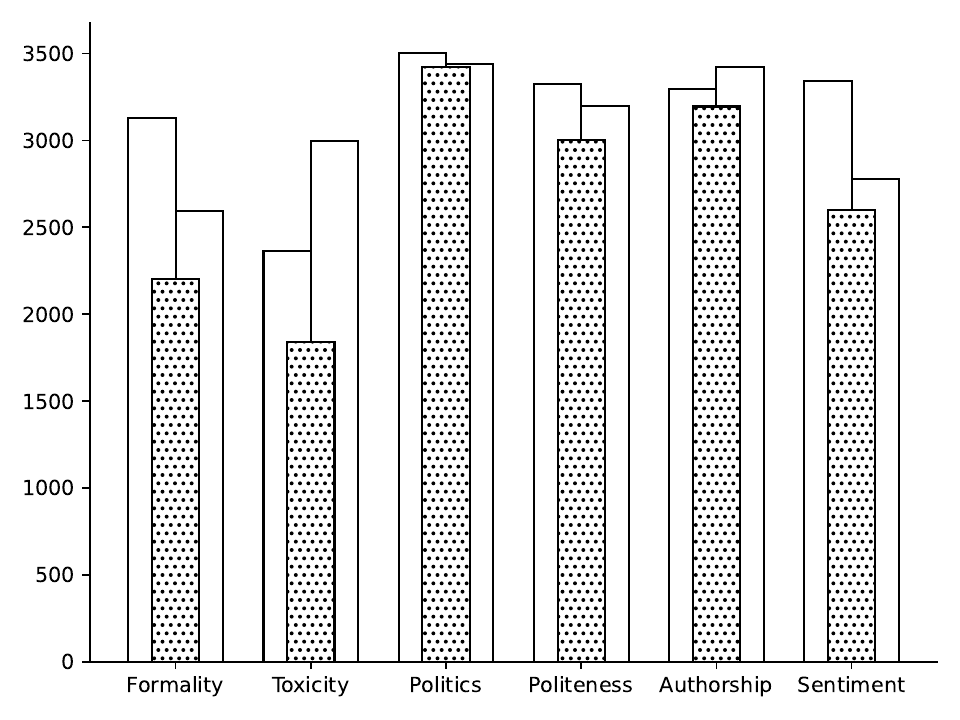}\vspace{-0.5em}
	\caption{\label{fig:overlap} Overlap statistics of style-specific neurons identified using the method of~\cite{tang2024language} on six benchmarks.
 }
 \vspace{-1.2em}
\end{figure}

\subsubsection{Neuron Selection}
Recently,~\citet{tang2024language} introduced a method for identifying language-specific neurons and demonstrated a significant overlap among neurons across different languages, such as an approximate 25\% overlap between Chinese and English neurons.
However, their study did not evaluate the performance implications of these overlaps.
We measure the overlap of style-specific neurons by applying the method of~\citet{tang2024language} directly to a style-specific corpus.
As illustrated in Figure~\ref{fig:overlap}, we observe a higher overlap among style-specific neurons.
For instance, in the Politics benchmark, nearly 95\% of neurons overlap between ``democratic'' and ``republican'' styles.
Moreover, we demonstrate that this substantial overlap negatively impacts the performance of TST (Section~\ref{sec:ablation}).

To eliminate the overlap between neurons of different styles, we identify style-specific neurons and their intersection.
Formally, suppose we have two distinct styles, denoted as $A$ and $B$.
We feed the corpora of the two styles to an LLM separately, to obtain the activation values of the neurons in the FFN layers for both styles, as described in Eq~(\ref{eq:act_val}).
We then select the neurons whose activation value exceeds zero, forming two sets denoted as $S_A$ and $S_B$, respectively.
Subsequently, we sort the activation values within $S_A$ and $S_B$ in descending order and select the neurons with the top $k$ values ($k = 500n, n\in\{1,2,3,\dots,20\}$ tuned on the validation set), resulting in $S_{A}^{\prime}$ and $S_{B}^{\prime}$.
Finally, we identify the neurons associated with strictly one of the styles by computing the disjoint sets of the two smaller sets:
$N_A = S_{A}^{\prime} \setminus S_{B}^{\prime}$ and $N_B = S_{B}^{\prime} \setminus S_{A}^{\prime}$.

\subsection{Deactivating Source Style Neurons}
\label{sec:deact}
After identifying neurons associated with a particular style, a common practice~\cite{tang2024language} is to deactivate these neurons by setting their activation values to zero during the model's forward pass.
However, neurons are sensitive components in neural networks; thus, deactivating a neuron associated with a specific feature (e.g., formal style) can lead to significant performance deterioration~\cite{morcos2018blog}.
To investigate the effects of deactivating source- and target-style neurons in TST task, we conduct experiments focusing on formality and politeness transfer tasks.

%% add the experiments
\begin{table}[t]
\centering
% \begin{subtable}{\columnwidth}
    \centering
    \resizebox{\columnwidth}{!}{
        \begin{tabular}{cc|cccc}
        \toprule
        \multicolumn{6}{c}{\textbf{Style Accuracy}} \\
        \midrule
        \multirow{3}{*}{\textbf{Source}} & \multirow{3}{*}{\textbf{Target}} & \multicolumn{2}{c}{\textbf{Formality}} & \multicolumn{2}{c}{\textbf{Politeness}} \\
        \cmidrule(lr){3-4}\cmidrule(lr){5-6}
        &                         & informal       & formal       & impolite       & polite       \\
        \midrule
        \ding{55}                & \ding{55}                & 80.00 & 11.20 & 79.50 & 14.80         \\
        \ding{51}                  & \ding{55}                & \textbf{80.53} & \textbf{13.63} & \textbf{80.06} & \textbf{19.37}        \\
        \ding{55}                & \ding{51}                  & 76.25 & 8.51  & 65.50 & 9.27        \\
        \ding{51}                  & \ding{51}                  & 78.42 & 9.27  & 73.48 & 10.36     \\
        \midrule
        \multicolumn{6}{c}{\textbf{Fluency}} \\
        \midrule
        \multirow{3}{*}{\textbf{Source}} & \multirow{3}{*}{\textbf{Target}} & \multicolumn{2}{c}{\textbf{Formality}} & \multicolumn{2}{c}{\textbf{Politeness}} \\
        \cmidrule(lr){3-4}\cmidrule(lr){5-6}
        &                         & informal       & formal       & impolite       & polite       \\
        \midrule
        \ding{55}                & \ding{55}                & \textbf{92.53}  & \textbf{87.69}  & \textbf{105.35} & \textbf{92.34}         \\
        \ding{51}                  & \ding{55}              & 104.17 & 96.83  & 127.26 & 105.12       \\
        \ding{55}                & \ding{51}                & 113.14 & 106.23 & 136.10 & 112.51        \\
        \ding{51}                  & \ding{51}              & 108.22 & 100.79 & 131.22 & 108.64     \\
        \bottomrule
        \end{tabular}
        }
\caption{
\label{tab:deact_src_tgt}
Experiments for deactivating neurons on formality and politeness benchmarks.
\ding{51} means the neuron is deactivated, while \ding{55} means the neuron is activated.
``Source'' and ``Target'' denotes the neuron sides.
The indicated style (e.g. formal) within a task (e.g. Formality) indicates the source, and its pair is the target style.
Style accuracy and fluency are defined in Section~\ref{sec:eval_metric}.
}
\vspace{-1.2em}
\end{table}

From Table~\ref{tab:deact_src_tgt}, we observe that:
\textbf{(1)} Deactivating the source-style neurons while keeping the target-style neurons active improves the accuracy of generating the target style.
Conversely, deactivating the target-style neurons, regardless of the state of the source-style neurons, leads to a decrease in the accuracy of generating the target style.
This occurs because deactivating the target-style neurons impairs the ability of LLMs to generate target-style words during decoding, resulting in lower accuracy.
On the other hand, deactivating the source-style neurons allows LLMs to focus more on generating target-style words, thus improving target style accuracy.
This finding aligns with related work on language-specific neuron deactivation~\cite{tang2024language,zhao2024large}.
\textbf{(2)} Fluency decreases whenever neurons are deactivated, whether they are source-style or target-style neurons.
This is mainly due to the significant impact that deactivating neurons has on the word distribution during decoding.
Specifically, the model tends to generate words of the non-deactivated style with a higher probability, leading to generated texts that are simply a concatenation of non-deactivated style words, thereby compromising fluency.
As illustrated in Figure~\ref{fig:framework}, after deactivating the source-style neurons, the generated text includes both ``Neither'' and ``quality''— two target-style words without maintaining sentence fluency.

\subsection{Contrastive Decoding for TST}
\label{sec:contras_decode}
Contrastive decoding (CD;~\citealp{li-etal-2023-contrastive}), which adjusts the probability of predicting the next word by comparing the outputs of a LLM with a weaker, smaller model, has been proven effective in enhancing fluency and coherence.
More recently,~\citet{chuang2024dola} proposed Dola, a CD approach that achieves excellent results by comparing outputs between the final layer and the early layers.
We adapt Dola to TST to mitigate the fluency issues observed during neuron deactivation.

\subsubsection{Dola}
Given a sequence of tokens $\{x_1, x_2, \dots, x_{t-1}\}$ and the total number ($N$) of layers in LLMs, the probability of the next token $x_t$ in $j$-th transformer layer can be computed in advance (known as \textit{early exit};~\citealp{schuster2022confident}) as:
% \vspace{-0.2em}
\begin{equation}
    % \vspace{-0.2em}
    \begin{aligned}
        p^j(x_{t} \mid x_{<t}) = \mathrm{softmax}\bigl(\phi(h_t^{(j)})\bigr)_{x_{t}}
    \end{aligned}
\end{equation}
where $h_t$ is the hidden states obtained from the embedding layer. $\phi(\cdot)$ is the vocabulary head used to predict the probabilities of the tokens.

Dola aims to contrast the information of the final layer and a set of early layers ($\mathcal{J} \subset \{0, \dots, N-1\}$) to obtain the next-token probability as:
% \vspace{-0.2em}
% \begin{ourfont}
%     \begin{equation}\label{eq:contras_decode}
%     % \vspace{-0.2em}
%     \begin{aligned}
%         \hat{p}(x_{t} \mid x_{<t}) = \mathrm{softmax}\bigl(\mathcal{F}\bigl(p^N(x_{t}), p^M(x_{t})\bigr)\bigr)_{x_t}
%     \end{aligned}
% \end{equation}
% \end{ourfont}

\begin{small}
    \begin{equation}\label{eq:contras_decode}
    % \vspace{-0.2em}
    \begin{aligned}
        \hat{p}(x_{t} \mid x_{<t}) = \mathrm{softmax}\bigl(\mathcal{F}\bigl(p^N(x_{t}), p^M(x_{t})\bigr)\bigr)_{x_t}
    \end{aligned}
\end{equation}
\end{small}
    
where $\mathcal{F}(\cdot)$ is the function used to contrast between the output distributions from one premature layer $M$ and the final layer by computing the log-domain difference between two distributions~\cite{li-etal-2023-contrastive} as follows:
% \vspace{-0.8em}

\begin{small}
    \begin{equation}
    % \vspace{-0.35em}
    \begin{aligned}
        \mathcal{F}\bigl(p^N(x_{t}), p^M(x_{t})\bigr) = \begin{cases} \log \dfrac{p^N(x_{t})}{p^M(x_{t})}, & \text { if } x_t\in\Phi, \\
-\infty , & \text { otherwise. }\end{cases} 
    \end{aligned}
\end{equation}
\end{small}

\noindent
% \textcolor{red}{where the subset $\mathcal{V}_{\text {head }}$ is defined as whether or not the token has high enough output probabilities from the mature layer.}
where $\Phi$ is defined as whether or not the token has high enough output probabilities from the mature layer as: 
% \vspace{-1.5em}
\begin{equation}
% \vspace{-0.65em}
    \Phi\left(x_{t} \mid x_{<t}\right)=\left\{p^{N}\left(x_{t}\right) \geq \max _{w} p^{N}(w)\right\}
\end{equation}
Layer $M$, the \textit{premature layer}, is selected dynamically at each time step by taking the layer with the largest Jensen-Shannon Divergence (JSD;~\citealp{menendez1997jensen}) to contrast output distributions from the final and the set of early candidate layers.
% $\mathcal{J}$.

\subsubsection{Our adaptation to TST}
\label{sec:our_adapt}
\begin{figure}[thbp]
      \centering
	   \begin{subfigure}{0.49\linewidth}
		\includegraphics[width=\linewidth]{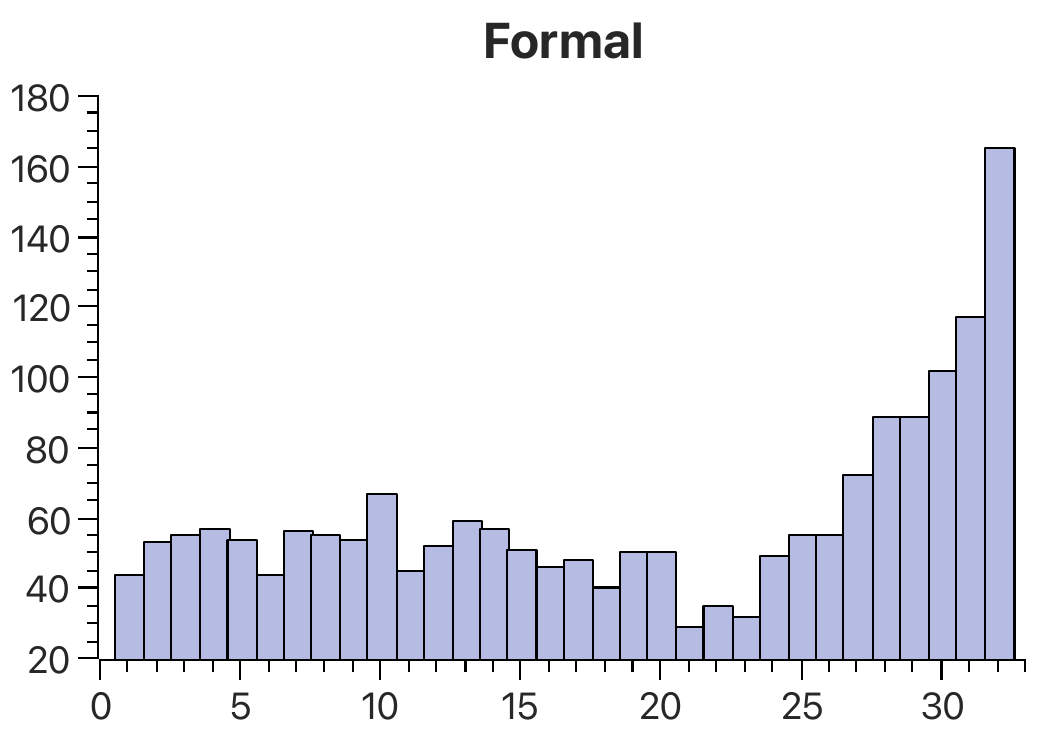}
	   \end{subfigure}
	   \begin{subfigure}{0.49\linewidth}
		\includegraphics[width=\linewidth]{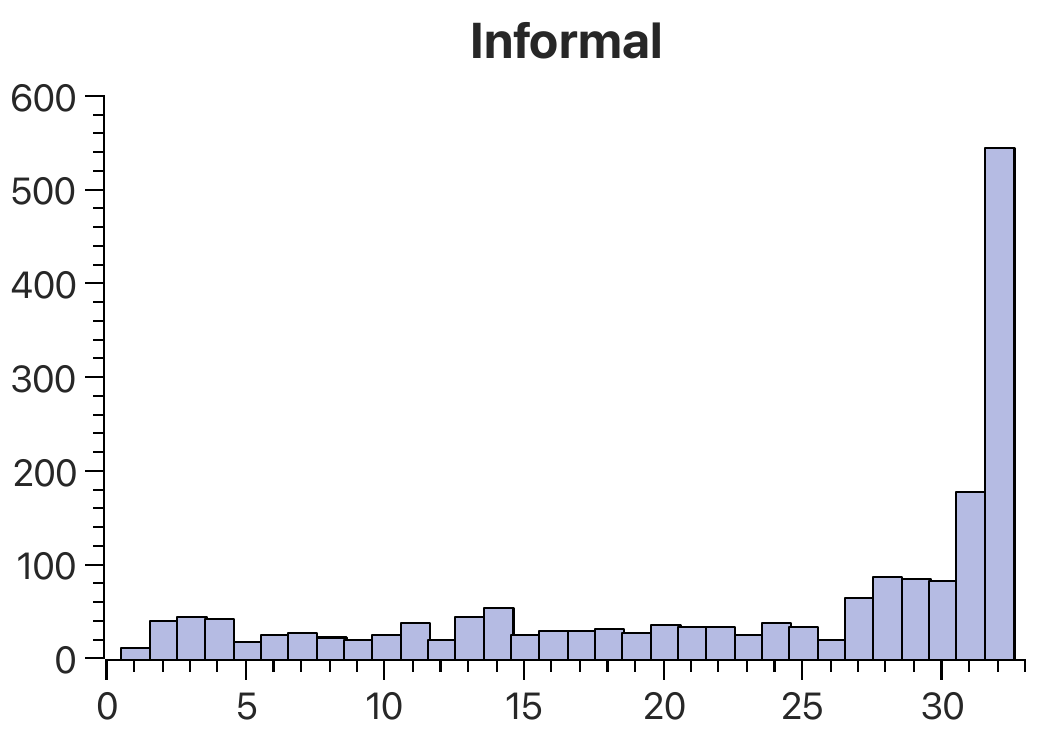}
	    \end{subfigure}
	     \begin{subfigure}{0.49\linewidth}
		 \includegraphics[width=\linewidth]{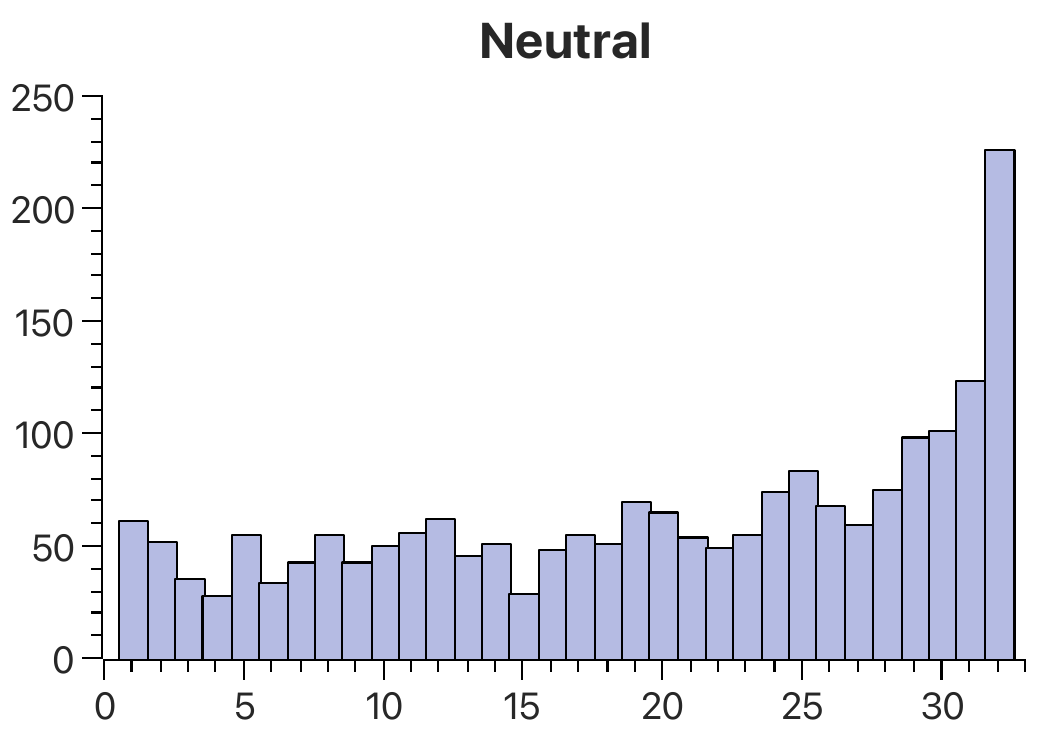}
	      \end{subfigure}
	       \begin{subfigure}{0.49\linewidth}
		  \includegraphics[width=\linewidth]{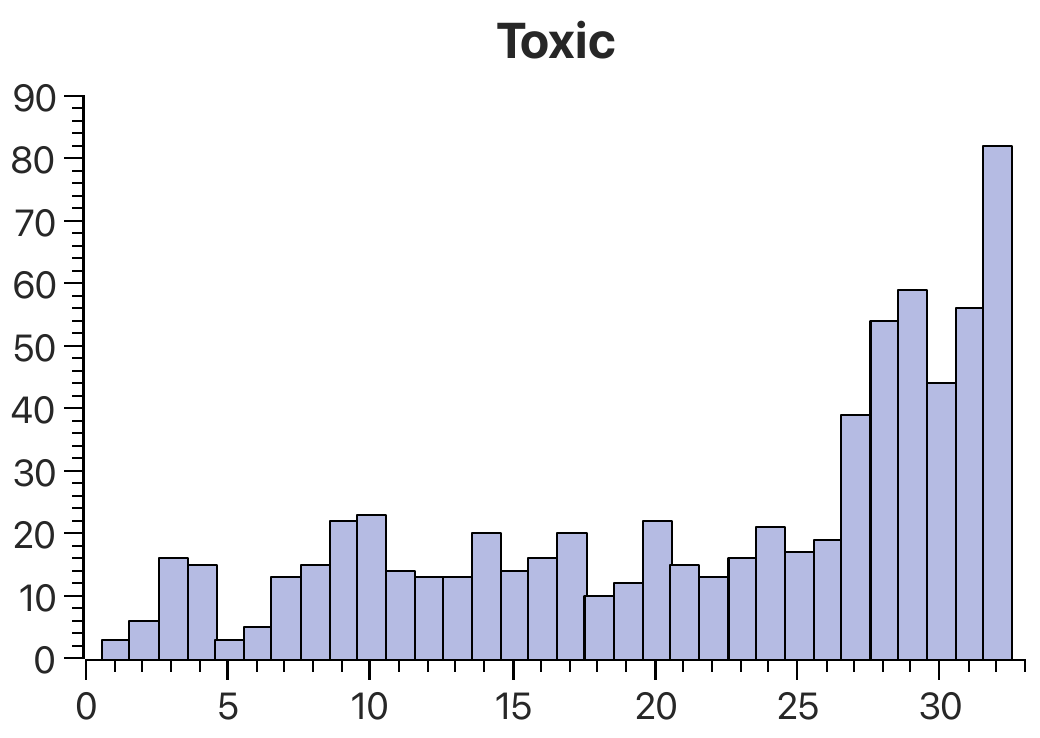}
	       \end{subfigure}
	\caption{Statistics of the number of style-specific neurons in each layer in LLaMA-3 on formality and toxicity benchmarks.}
	\label{fig:neu_dis}
 \vspace{-1.5em}
\end{figure}
\paragraph{Candidate layer selection.}
To better adapt Dola to TST, we select candidate layers for comparison based on the criterion that these layers should contain more style information.
To this end, we measure the amount of style-specific neurons across each layer.
As shown in Figure~\ref{fig:neu_dis}, the last few layers, particularly the final layer, contain significantly more style neurons compared to the earlier layers.
Therefore, we select the last few layers ($4$ in our experiments) as our candidate layers.

\paragraph{Next-token prediction.}
After deactivating the source-style neurons, LLMs tend to generate target-style tokens.
However, we need to determine whether the appearance of these target-style tokens is due to their consistently high probability from the early layers to the final layer or due to a probability shift caused by neuron deactivation in the last few layers.
If the probability of tokens at a given time step remains consistent from the first layer to the final layer, it indicates that these tokens are style-independent (typically function words) and are retained in the output of the final layer by Eq.~(\ref{eq:contras_decode}).
Conversely, if these words have a low probability in the early layers (typically target-style words) and only exhibit a probability ``mutation'' in the last few layers due to the deactivation of source-style neurons, we then select the layer with the maximum JSD distance from the candidate layers as our premature layer $M$ and adjust their probability distribution according to Eq.~(\ref{eq:contras_decode}).
\section{Experiments}
\label{sec:exp}

\subsection{Datasets}
\label{sec:datasets}
We evaluate our approach on six typical TST tasks: formality, toxicity, politics, politeness, authorship, and sentiment on GYAFC~\cite{rao-tetreault-2018-dear}, ParaDetox~\cite{logacheva-etal-2022-paradetox}, Politeness~\cite{madaan-etal-2020-politeness}, Shakespeare~\cite{xu-etal-2012-paraphrasing} and Yelp~\cite{shen2017style}.
The statistics of the datasets can be found in Appendix~\ref{app:dataset}.

\subsection{Baselines}
We compare our approach with the following baselines:
(1) \textbf{LLaMA-3:} We use LLaMA-3~\cite{llama3} without additional fine-tuning as the vanilla baseline system. 
(2) \textbf{APE:} Using activation probability entropy to identify the style specific neurons~\cite{tang2024language}.
(3) \textbf{AVF:} Using activation value frequency and set a threshold to identify the style neurons~\cite{tan2024neuron}. 
(4) \textbf{PNMA:} Finding neurons that activate on the source style sentences but do not activate on target style sentences~\cite{kojima2024multilingual}.
Note that (2), (3), and (4) from the original paper focus on identifying language-specific neurons to enhance the multilingual capabilities of LLMs, and we extend these methodologies to our style-related corpus.
For (4), it requires the use of parallel data from both source and target texts to identify neurons, whereas (2), (3), and our method does not require the use of parallel data.
Additionally, after identifying the neurons, we deactivate the source-style neurons in (2), (3), and (4).
For a detailed comparison of various decoding strategies, please refer to Appendix~\ref{app:diff_decoding}.

\subsection{Implementation}
We use the 8B model of LLaMA-3, available in the HuggingFace repository\footnote{\url{https://github.com/huggingface/transformers}} in zero-shot setting.
To further assess the scalability of our method, we also employ the 70B LLaMA-3 model (Appendix~\ref{app:model_size}).
For each baseline system, we use the same hyperparameters (e.g., threshold) as the original paper.

\subsection{Evaluation Metric}
\label{sec:eval_metric}
We evaluate our approach using three metrics commonly employed in TST tasks.
\textbf{Style Accuracy.}
Accuracy of labels predicted as correct by a style classifier. Please refer to Appendix~\ref{app:cls} for more details.
\textbf{Content Preservation.}
Cosine similarity between the embeddings of the original text and the text generated by the model, using LaBSE~\cite{feng-etal-2022-language} to obtain sentence embeddings as our primary metric.
Additionally, we employ BLEURT metrics~\cite{sellam-etal-2020-bleurt} for comparison, as recent studies indicate strong correlations between BLEURT assessments on TST and human evaluation results (Appendix~\ref{app:diff_content_prev}).
\textbf{Fluency.}
Perplexity of the generated sentences using GPT-2~\cite{radford2019language}.
\section{Results}
\label{sec:res}
%% Table: main results
\begin{table*}[t]
\resizebox{\textwidth}{!}{
\begin{tabular}{l|cccccccccccc}
\toprule
\multicolumn{13}{c}{\textbf{Style Transfer Accuracy}} \\
\midrule
        & \multicolumn{2}{c}{\textbf{Formality}} & \multicolumn{2}{c}{\textbf{Toxicity}} & \multicolumn{2}{c}{\textbf{Politics}} & \multicolumn{2}{c}{\textbf{Politeness}} & \multicolumn{2}{c}{\textbf{Authorship}} & \multicolumn{2}{c}{\textbf{Sentiment}} \\
\cmidrule(lr){2-3}\cmidrule(lr){4-5}\cmidrule(lr){6-7}\cmidrule(lr){8-9}\cmidrule(lr){10-11}\cmidrule(lr){12-13}
        & informal       & formal       & toxic         & neutral      & democratic    & republican   & impolite       & polite       & shakespeare    & modern        & positive       & negative     \\
\midrule
LLaMA-3 & 80.00          & 11.20        & 47.67         & 29.04        & 35.50         & 48.20        & 79.50          & 14.80        & 63.80          & 43.80         & 76.40          & 52.80        \\
APE    & 74.00          & 12.20        & 47.57         & 28.44        & \textbf{40.90}         & 44.80        & 77.10          & 18.20        & 55.80          & 44.60         & 78.90          & 48.00        \\
AVF    & 76.00          & 12.40        & 47.57         & 28.44        & 38.80         & 44.20        & 77.90          & 18.70        & 55.60          & 44.40         & \textbf{79.20}          & 47.90        \\
PNMA    & 73.85          & 8.70         & 42.43         & 23.79        & 35.57         & 37.05        & 72.84          & 14.16        & 53.74          & 37.58         & 75.39          & 41.71        \\
Our     & \textbf{80.80}          & \textbf{14.40}        & \textbf{55.36}         & \textbf{31.98}        & 37.81         & \textbf{50.30}        & \textbf{80.63}          & \textbf{23.27}        & \textbf{73.40}          & \textbf{45.14}         & 77.93          & \textbf{54.73}        \\
\midrule
\midrule
\multicolumn{13}{c}{\textbf{Content Preservation}} \\
\midrule
        & \multicolumn{2}{c}{\textbf{Formality}} & \multicolumn{2}{c}{\textbf{Toxicity}} & \multicolumn{2}{c}{\textbf{Politics}} & \multicolumn{2}{c}{\textbf{Politeness}} & \multicolumn{2}{c}{\textbf{Authorship}} & \multicolumn{2}{c}{\textbf{Sentiment}} \\
\cmidrule(lr){2-3}\cmidrule(lr){4-5}\cmidrule(lr){6-7}\cmidrule(lr){8-9}\cmidrule(lr){10-11}\cmidrule(lr){12-13}
        & informal       & formal       & toxic         & neutral      & democratic    & republican   & impolite       & polite       & shakespeare    & modern        & positive       & negative     \\
\midrule
LLaMA-3 & \textbf{85.95}          & 74.71        & 73.54         & 82.71        & 82.48         & 75.77        & 75.32          & \textbf{89.14}        & 78.75          & \textbf{62.28}         & 76.17          & \textbf{74.47}        \\
APE    & 76.72          & 85.06        & \textbf{76.72}         & 83.00        & \textbf{87.99}         & \textbf{82.21}        & 76.80          & 87.89        & 80.07          & 57.61         & \textbf{76.52}          & 73.53        \\
AVF    & 75.21          & 84.53        & 76.63         & \textbf{83.57}        & 86.92         & 80.68        & \textbf{76.94}          & 87.32        & \textbf{80.94}          & 58.98         & 76.15          & 73.95        \\
PNMA    & 75.52          & 84.11        & 75.67         & 82.54        & 86.79         & 80.67        & 76.04          & 86.93        & 79.22          & 57.42         & 75.04          & 72.67        \\
Our     & 85.84          & \textbf{86.28}        & 75.85         & 80.10        & 82.32         & 74.96        & 75.65          & 82.47        & 77.19          & 60.92         & 75.25          & 74.21        \\
\midrule
\midrule
\multicolumn{13}{c}{\textbf{Fluency}} \\
\midrule
        & \multicolumn{2}{c}{\textbf{Formality}} & \multicolumn{2}{c}{\textbf{Toxicity}} & \multicolumn{2}{c}{\textbf{Politics}} & \multicolumn{2}{c}{\textbf{Politeness}} & \multicolumn{2}{c}{\textbf{Authorship}} & \multicolumn{2}{c}{\textbf{Sentiment}} \\
\cmidrule(lr){2-3}\cmidrule(lr){4-5}\cmidrule(lr){6-7}\cmidrule(lr){8-9}\cmidrule(lr){10-11}\cmidrule(lr){12-13}
        & informal       & formal       & toxic         & neutral      & democratic    & republican   & impolite       & polite       & shakespeare    & modern        & positive       & negative     \\
\midrule
LLaMA-3 & 92.53          & 87.69        & 113.84        & 191.30       & 88.22         & 68.49        & 105.35         & 92.34        & 197.62         & 136.03        & 177.01         & 125.98       \\
APE    & 94.27          & 89.93        & 133.12        & 188.34       & 88.51         & 69.06        & 108.24         & 95.17        & 250.65         & 133.92        & \textbf{151.06}         & 126.73       \\
AVF    & 96.63          & 89.36        & 131.10        & 191.29       & 87.93         & 75.94        & 112.67         & 97.50        & 220.30          & \textbf{126.42}        & 151.33         & 130.17       \\
PNMA    & 103.61         & 90.85        & 136.27        & 194.71       & 96.31         & 77.95        & 111.77         & 101.61       & 260.52         & 135.00        & 154.85         & 129.49       \\
Our     & \textbf{90.79}          & \textbf{81.46}        & \textbf{85.65}         & \textbf{172.26}       & \textbf{85.28}         & \textbf{66.68}        & \textbf{104.92}         & \textbf{83.36}        & \textbf{151.71}         & 134.86        & 174.46         & \textbf{110.48}       \\
\bottomrule
\end{tabular}
}
\caption{
\label{tab:main_res}
\textbf{Main Results:} Style transfer accuracy (higher values are better; $\uparrow$), content preservation ($\uparrow$) and fluency ($\downarrow$) on $6$ datasets across $12$ transfer directions.
Best results are highlighted in bold.
}
\vspace{-0.5em}
\end{table*}
Table~\ref{tab:main_res} shows the transfer performance (style accuracy, content preservation and fluency) of the six benchmarks in $12$ directions.

\textbf{Overall Performance.}
While the \emph{APE}, \emph{AVF}, and \emph{PNMA} demonstrate strong performance in enhancing multilingual capabilities, they do not outperform the original LLaMA-3 model in the TST task, with the exception of the content preservation metric.
This disparity arises primarily because language-specific properties can be identified using straightforward features, such as script differences.
Consequently, the neuron selection methods of these baselines, despite their partial overlaps, have minimal impact on multilingual performance.
However, text style represents a more complex attribute, requiring models to learn extensive knowledge and execute nuanced judgments at both the word and semantic levels.
The overlap of neurons in baseline systems across source and target styles adversely affects the results, particularly in style accuracy.
Furthermore, the baseline methods lack a contrastive decoding strategy, which compromises their fluency.
Our method outperforms the baseline methods in terms of both accuracy and fluency, highliting the importance of eliminating overlapping style neurons and employing contrastive decoding.

\textbf{Content Preservation.}
Interestingly, we observe that the original LLaMA-3 and other baseline systems exhibit strong performance in content preservation, which appears inconsistent with conclusions drawn from the other two metrics.
Upon closer examination, we find that this content preservation is largely attributable to the copy mechanism, i.e., the generated text tends to prioritize maintaining the original semantics, thereby neglecting the stylistic differences.
A detailed discussion on this can be found in Section~\ref{sec:copy}.
Another potential explanation is the semantic gap, which varies significantly between sentences of different styles, and for which no effective metric currently exists to fully measure this gap.
For example, when transferring text from an informal to a formal style, the original text \emph{``Sorry about that.''} and the target text \emph{``I apologize for the inconvenience caused.''} are stylistically aligned, but they diverge significantly in semantic space.
This is reflected in a low cosine similarity score of $0.447$ between them.

\textbf{Different Directions.}
We observe significant performance discrepancies when transferring between different directions within the same task.
For example, transferring from impolite to polite achieves a style accuracy of nearly 80\%, whereas the reverse direction achieves only about 12\%.
This disparity can be attributed to the training data of LLMs, which predominantly consist of positive corpora (e.g., polite, neutral, formal), with inadequate representation from negative corpora.
Additionally, LLMs have a tendency to generate safer responses~\cite{touvron2023llama}, which can compromise the utility of tasks involving style transfer.
\section{Analysis}
\label{sec:analysis}
In this section, we conduct an ablation study to verify the criticality of eliminating overlap between source- and target-side style neurons, alongside the importance of neuron deactivation and contrastive decofing (Section~\ref{sec:ablation}).
Subsequently, we conduct a detailed analysis of the copy problem in the TST task (Section~\ref{sec:copy}). Finally, we delve into several other significant factors inherent to our approach (Section~\ref{sec:further}).

%% overlap ablation
\begin{table}[t]
\resizebox{\columnwidth}{!}{
\begin{tabular}{ll|cc}
\toprule
\multicolumn{2}{c}{\textbf{Style}}  & \textbf{without} & \textbf{with}  \\
\midrule
\multirow{2}{*}{\textbf{Formality}}  & informal$\rightarrow$formal       & 74.00   & \textbf{79.40} \\
                            & formal$\rightarrow$informal       & 12.20   & \textbf{13.63} \\
\midrule
\multirow{2}{*}{\textbf{Toxicity}}   & toxic$\rightarrow$neutral         & 47.57   & \textbf{49.78} \\
                            & neutral$\rightarrow$toxic         & 28.44   & \textbf{29.82} \\
\midrule
\multirow{2}{*}{\textbf{Politics}}   & democratic$\rightarrow$republican & \textbf{40.90}   & 37.51 \\
                            & republican$\rightarrow$democratic & 44.80   & \textbf{49.70} \\
\midrule
\multirow{2}{*}{\textbf{Politeness}} & impolite$\rightarrow$polite       & 77.10   & \textbf{80.10} \\
                            & polite$\rightarrow$impolite       & 18.20   & \textbf{21.73} \\
\midrule
\multirow{2}{*}{\textbf{Authorship}} & shakespeare$\rightarrow$modern    & 55.80   & \textbf{63.00} \\
                            & modern$\rightarrow$shakespeare    & 44.60   & \textbf{45.42} \\
\midrule
\multirow{2}{*}{\textbf{Sentiment}}  & positive$\rightarrow$negative     & 78.90   & \textbf{79.75} \\
                            & negative$\rightarrow$positive     & 48.00   & \textbf{51.70} \\
\bottomrule
\end{tabular}}
\caption{
\label{tab:ablation_overlap}
\textbf{Ablation study:} Style transfer accuracy on removing overlap between source- and target-side style neurons in six benchmarks.
``with'' indicates the removal of overlap.
}
\vspace{-1.2em}
\end{table}

\subsection{Ablation Study}
\label{sec:ablation}
We conduct an ablation study, detailed in Table~\ref{tab:ablation_overlap}, to evaluate the effectiveness of removing overlapping source- and target-style neurons.
The results demonstrate a considerable advantage in eliminating such overlap compared to allowing mixed patterns of neuron activation.
As highlighted by the statistics in Section~\ref{sec:identify}, there is a substantial 95\% overlap in most neurons, indicating that source style neurons largely coincide with target style neurons, meking them nearly indistinguishable when directly decoding using LLMs.

%%% ablation:deact_dola
\begin{table}[t]
\resizebox{\columnwidth}{!}{
\begin{tabular}{l|cccccc}
\toprule
    & \multirow{3}{*}{\textbf{Deactivate}} & \multirow{3}{*}{\textbf{Contrastive}} & \multicolumn{2}{c}{\textbf{Toxicity}} & \multicolumn{2}{c}{\textbf{Authorship}} \\
\cmidrule(lr){4-5}\cmidrule(lr){6-7}
    &                         &                         & toxic       & neutral       & shakespeare       & modern       \\
    % &                         &                         & $\rightarrow$  & $\leftarrow$ & $\rightarrow$  & $\leftarrow$ \\
\midrule
\#1 & \ding{55}                & \ding{55}                & 47.67          & 29.04        & 63.80           & 43.80         \\
\#2 & \ding{51}                  & \ding{55}                & 52.63          & 31.07        & 68.39          & 44.71        \\
\#3 & \ding{55}                & \ding{51}                  & 46.82          & 28.31        & 63.23          & 43.16        \\
\#4 & \ding{51}                  & \ding{51}                  & \textbf{55.36}          & \textbf{31.98}        & \textbf{73.40}           & \textbf{45.14}       \\
\bottomrule
\end{tabular}
}
\caption{
\label{tab:ablation_decoding}
\textbf{Ablation study:} Style transfer accuracy for neuron deactivation and contrastive decoding on the toxicity and authorship tasks. 
``\ding{51}'' means the inclusion of the neuron deactivation or contrastive decoding steps, while ``\ding{55}'' means they are turned off.
\#1 indicates the results from baseline LLaMA-3 model, which do not use the deactivation nor the contrastive steps.
}
\vspace{-1.2em}
\end{table}
Additionally, Table~\ref{tab:ablation_decoding} presents the results of ablating neuron deactivation and contrastive decoding (CD).
Our findings are as follows:
\textbf{(1)}~Comparing \#1 and \#2, we observe a significant impact of deactivating neurons on the final results. 
This is because deactivating neurons on the source side encourages the LLMs to generate words in the target style. 
\textbf{(2)}~Comparing \#1 and \#3, we find that using CD alone does not significantly improve and may even degrade the results.
This is attributed to the fact that style-related information is processed in later layers, and simply comparing these layers does not yield substantial improvements.
Without deactivating neurons, the target style words are not effectively generated, resulting in minimal JSD distance between the style layers and the final layer, thereby reducing the effectiveness of CD.
\textbf{(3)}~Experiment \#4 demonstrates that optimal performance is achieved when both deactivating source-side style neurons and employing CD.
Deactivating neurons enhances the probability to generate target style vocabulary, as discussed in Section~\ref{sec:deact}, albeit at the cost of fluency in generated sentences.
Therefore, CD proves crucial in further enhancing the fluency of sentences.

\subsection{Copy Problem}
\label{sec:copy}
\begin{figure}[!t]
\centering
	\includegraphics[width=\linewidth]{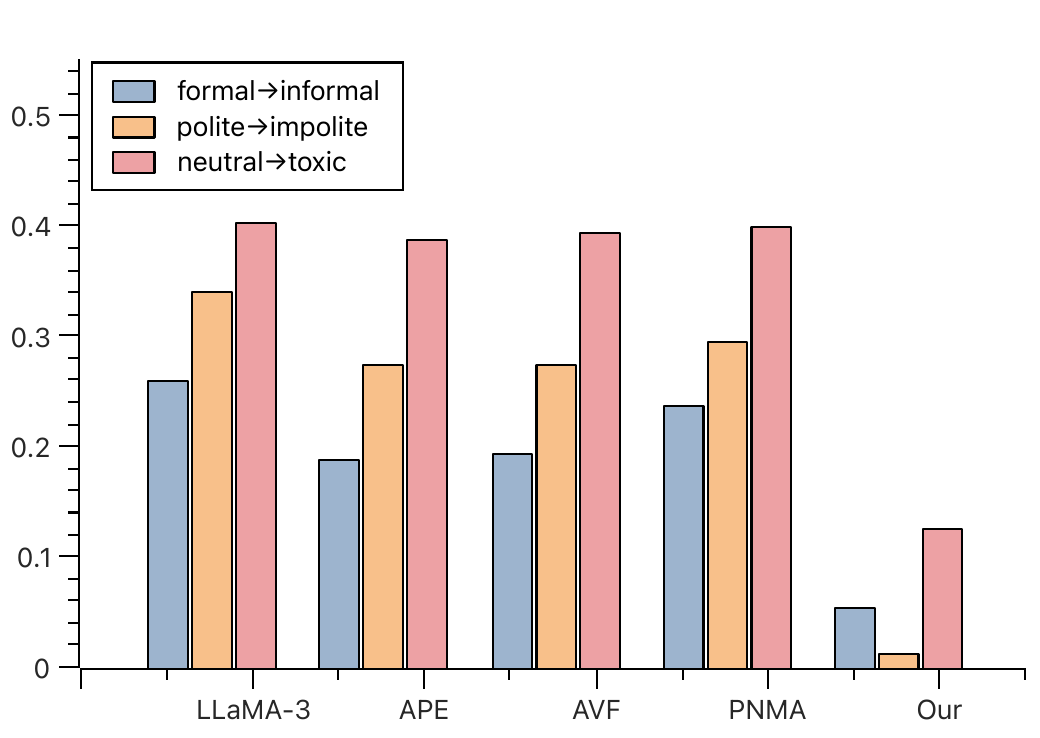}
	\caption{\label{fig:copy} Copy Ratio on three selected TST tasks. Lower value indicates better performance of the model.
 }
 \vspace{-1.2em}
\end{figure}

The ``copy problem'' arises when models simply reproduce the input text unchanged in the output, a challenge prevalent in multilingual machine translation~\cite{lai-etal-2023-mitigating,lai2023extending}.
Given the goal to maintain semantic consistency of the input sentences in TST, LLMs often resort to direct copying.
To investigate this phenomenon, we analyze tasks related to \emph{formality}, \emph{politeness}, and \emph{toxicity}.
Figure~\ref{fig:copy} illustrates a significant number of copy instances in the original LLaMA-3, indicating a preference for preserving semantic meaning rather than incorporating stylistic variations in TST.
Neuron-based approaches (\emph{APE}, \emph{AVF}, and \emph{PNMA}) partially mitigate this issue by controlling neuron activation, thereby producing more target-style words during decoding, as evidenced in Section~\ref{tab:ablation_overlap}.
However, these baselines suffer in performance due to their inability to fully eliminate overlap between source and target style neurons.
In contrast, our approach achieves a reduced copy rate by deactivating source-side neurons and employing a novel decoding strategy.

\vspace{-0.2em}

\subsection{Case Study}
\label{sec:case}
%%%
%% case study
\begin{table*}[t]
\centering
\begin{tabular}{c}
\includegraphics[width=\linewidth]{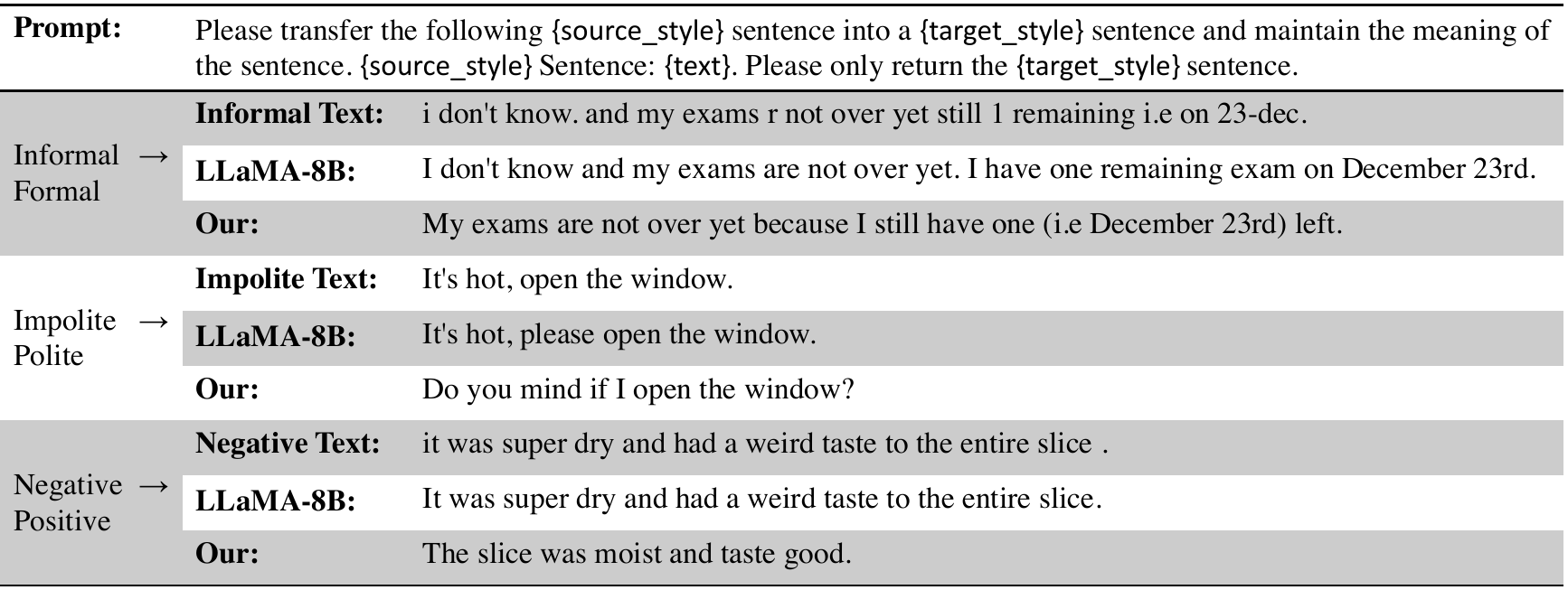}
\end{tabular}
\caption{Case study on informal$\rightarrow$formal, impolite$\rightarrow$polite and negative$\rightarrow$positive tasks.}
\label{tab:case_study}
\end{table*}
%%%
To further demonstrate the effectiveness of our approach, we conduct a case study on three style transfer tasks: informal to formal, impolite to polite, and negative to positive.
As shown in Table~\ref{tab:case_study}, the original LLaMA model often retained a higher number of words from the source text, sometimes copying them verbatim—an issue discussed in Section~\ref{sec:copy}.
In contrast, by adjusting style neurons, we guided the model to generate more varied vocabulary aligned with the target style.
For instance, the model produced the word ``moist" as part of a positive style transformation.

\subsection{Further Analysis}
\label{sec:further}
We conduct a comprehensive analysis of our method across various dimensions, including different model (Appendix~\ref{app:model_size}), layer selection strategies (Appendix~\ref{app:style_layer}), content preservation metrics (Appendix~\ref{app:diff_content_prev}), and decoding strategies (Appendix~\ref{app:diff_decoding}), yielding several key insights:
\textbf{(1)}~Our method consistently demonstrates effectiveness across diverse model sizes, including larger models like 70B.
\textbf{(2)}~Selecting the last few layers proves optimal compared to earlier layers.
\textbf{(3)}~Different strategies for preserving meaning yield similar outcomes, highlighting the importance of exploring innovative approaches in future research.
\textbf{(4)}~Contrastive decoding exhibits significant advantages over traditional decoding methods in the TST task, motivating our adoption of CD strategy.
\section{Conclusion}
\label{sec:conclusion}
We revisit the TST task in LLMs through a neuronal analysis perspective.
Our study focuses on identifying style-specific neurons within LLMs, highlighting the critical importance of removing overlap between source- and target-side stylistic neurons.
We find that deactivating source-specific neurons enhances the probability of generating target-style words but may compromise the fluency of generated sentences.
To mitigate this issue, we adapt the state-of-the-art contrastive decoding method (Dola) for TST, ensuring both the fluency and effective style transformation of generated sentences.
Experimental results across six benchmarks demonstrate the efficacy of our approach.
\section{Limitations}
This work has the following limitations:
(1) We deactivate style-specific neurons across all layers; however, considering other layers may yield additional insights. For instance,~\citet{zhao2024large} found that deactivating neurons in different layers (e.g., understanding layer or generating layer) can have subtle effects on experimental results.
We will consider this as a direction for future research.
(2) We evaluate our approach only on the text style transfer task; however, our method has the potential to be applied to other style-related tasks, such as image style transfer~\cite{wang2024multimodality} and multilingual style transfer~\cite{mukherjee2024multilingual}.
Furthermore, our approach is task-agnostic, with significant potential to adapt to other tasks, such as identifying domain-specific neurons and applying them to domain adaptation tasks~\cite{lai-etal-2022-m4,lai-etal-2022-improving-domain}.

\section*{Acknowledgement}
The work was supported by the European Research Council (ERC) under the European Union’s Horizon Europe research and innovation programme (grant agreement No. 101113091) and by the German Research Foundation (DFG; grant FR 2829/7-1).


\newpage
\bibliography{custom,acl}

\begin{thebibliography}{51}
\providecommand{\natexlab}[1]{#1}

\bibitem[{Chang et~al.(2024)Chang, Wang, Wang, Wu, Yang, Zhu, Chen, Yi, Wang, Wang et~al.}]{chang2024survey}
Yupeng Chang, Xu~Wang, Jindong Wang, Yuan Wu, Linyi Yang, Kaijie Zhu, Hao Chen, Xiaoyuan Yi, Cunxiang Wang, Yidong Wang, et~al. 2024.
\newblock A survey on evaluation of large language models.
\newblock \emph{ACM Transactions on Intelligent Systems and Technology}, 15(3):1--45.

\bibitem[{Chen(2024)}]{chen2024lmstyle}
Jianlin Chen. 2024.
\newblock Lmstyle benchmark: Evaluating text style transfer for chatbots.
\newblock \emph{arXiv preprint arXiv:2403.08943}.

\bibitem[{Chen et~al.(2023)Chen, Zhao, Yu, McKeown, and He}]{chen-etal-2023-relation}
Yanda Chen, Chen Zhao, Zhou Yu, Kathleen McKeown, and He~He. 2023.
\newblock \href {https://doi.org/10.18653/v1/2023.findings-emnlp.12} {On the relation between sensitivity and accuracy in in-context learning}.
\newblock In \emph{Findings of the Association for Computational Linguistics: EMNLP 2023}, pages 155--167, Singapore. Association for Computational Linguistics.

\bibitem[{Chuang et~al.(2024)Chuang, Xie, Luo, Kim, Glass, and He}]{chuang2024dola}
Yung-Sung Chuang, Yujia Xie, Hongyin Luo, Yoon Kim, James~R Glass, and Pengcheng He. 2024.
\newblock Dola: Decoding by contrasting layers improves factuality in large language models.
\newblock In \emph{The Twelfth International Conference on Learning Representations}.

\bibitem[{Dementieva et~al.(2023)Dementieva, Moskovskiy, Dale, and Panchenko}]{dementieva-etal-2023-exploring}
Daryna Dementieva, Daniil Moskovskiy, David Dale, and Alexander Panchenko. 2023.
\newblock \href {https://doi.org/10.18653/v1/2023.ijcnlp-main.70} {Exploring methods for cross-lingual text style transfer: The case of text detoxification}.
\newblock In \emph{Proceedings of the 13th International Joint Conference on Natural Language Processing and the 3rd Conference of the Asia-Pacific Chapter of the Association for Computational Linguistics (Volume 1: Long Papers)}, pages 1083--1101, Nusa Dua, Bali. Association for Computational Linguistics.

\bibitem[{Dreyer et~al.(2024)Dreyer, Purelku, Vielhaben, Samek, and Lapuschkin}]{dreyer2024pure}
Maximilian Dreyer, Erblina Purelku, Johanna Vielhaben, Wojciech Samek, and Sebastian Lapuschkin. 2024.
\newblock Pure: Turning polysemantic neurons into pure features by identifying relevant circuits.
\newblock \emph{arXiv preprint arXiv:2404.06453}.

\bibitem[{Feng et~al.(2022)Feng, Yang, Cer, Arivazhagan, and Wang}]{feng-etal-2022-language}
Fangxiaoyu Feng, Yinfei Yang, Daniel Cer, Naveen Arivazhagan, and Wei Wang. 2022.
\newblock \href {https://doi.org/10.18653/v1/2022.acl-long.62} {Language-agnostic {BERT} sentence embedding}.
\newblock In \emph{Proceedings of the 60th Annual Meeting of the Association for Computational Linguistics (Volume 1: Long Papers)}, pages 878--891, Dublin, Ireland. Association for Computational Linguistics.

\bibitem[{Garde et~al.(2023)Garde, Kran, and Barez}]{garde2023deepdecipher}
Albert Garde, Esben Kran, and Fazl Barez. 2023.
\newblock Deepdecipher: Accessing and investigating neuron activation in large language models.
\newblock \emph{arXiv preprint arXiv:2310.01870}.

\bibitem[{Hu et~al.(2022)Hu, Lee, Aggarwal, and Zhang}]{hu2022text}
Zhiqiang Hu, Roy Ka-Wei Lee, Charu~C Aggarwal, and Aston Zhang. 2022.
\newblock Text style transfer: A review and experimental evaluation.
\newblock \emph{ACM SIGKDD Explorations Newsletter}, 24(1):14--45.

\bibitem[{Jin et~al.(2022)Jin, Jin, Hu, Vechtomova, and Mihalcea}]{jin-etal-2022-deep}
Di~Jin, Zhijing Jin, Zhiting Hu, Olga Vechtomova, and Rada Mihalcea. 2022.
\newblock \href {https://doi.org/10.1162/coli_a_00426} {Deep learning for text style transfer: A survey}.
\newblock \emph{Computational Linguistics}, 48(1):155--205.

\bibitem[{Kojima et~al.(2024)Kojima, Okimura, Iwasawa, Yanaka, and Matsuo}]{kojima2024multilingual}
Takeshi Kojima, Itsuki Okimura, Yusuke Iwasawa, Hitomi Yanaka, and Yutaka Matsuo. 2024.
\newblock On the multilingual ability of decoder-based pre-trained language models: Finding and controlling language-specific neurons.
\newblock \emph{arXiv preprint arXiv:2404.02431}.

\bibitem[{Lai et~al.(2022{\natexlab{a}})Lai, Chronopoulou, and Fraser}]{lai-etal-2022-m4}
Wen Lai, Alexandra Chronopoulou, and Alexander Fraser. 2022{\natexlab{a}}.
\newblock \href {https://doi.org/10.18653/v1/2022.findings-emnlp.315} {m$^4$ adapter: Multilingual multi-domain adaptation for machine translation with a meta-adapter}.
\newblock In \emph{Findings of the Association for Computational Linguistics: EMNLP 2022}, pages 4282--4296, Abu Dhabi, United Arab Emirates. Association for Computational Linguistics.

\bibitem[{Lai et~al.(2023{\natexlab{a}})Lai, Chronopoulou, and Fraser}]{lai-etal-2023-mitigating}
Wen Lai, Alexandra Chronopoulou, and Alexander Fraser. 2023{\natexlab{a}}.
\newblock \href {https://doi.org/10.18653/v1/2023.findings-emnlp.953} {Mitigating data imbalance and representation degeneration in multilingual machine translation}.
\newblock In \emph{Findings of the Association for Computational Linguistics: EMNLP 2023}, pages 14279--14294, Singapore. Association for Computational Linguistics.

\bibitem[{Lai et~al.(2023{\natexlab{b}})Lai, Hangya, and Fraser}]{lai2023extending}
Wen Lai, Viktor Hangya, and Alexander Fraser. 2023{\natexlab{b}}.
\newblock Extending multilingual machine translation through imitation learning.
\newblock \emph{arXiv preprint arXiv:2311.08538}.

\bibitem[{Lai et~al.(2022{\natexlab{b}})Lai, Libovick{\'y}, and Fraser}]{lai-etal-2022-improving-domain}
Wen Lai, Jind{\v{r}}ich Libovick{\'y}, and Alexander Fraser. 2022{\natexlab{b}}.
\newblock \href {https://aclanthology.org/2022.coling-1.461} {Improving both domain robustness and domain adaptability in machine translation}.
\newblock In \emph{Proceedings of the 29th International Conference on Computational Linguistics}, pages 5191--5204, Gyeongju, Republic of Korea. International Committee on Computational Linguistics.

\bibitem[{Li et~al.(2024)Li, Patel, Vi{\'e}gas, Pfister, and Wattenberg}]{li2024inference}
Kenneth Li, Oam Patel, Fernanda Vi{\'e}gas, Hanspeter Pfister, and Martin Wattenberg. 2024.
\newblock Inference-time intervention: Eliciting truthful answers from a language model.
\newblock \emph{Advances in Neural Information Processing Systems}, 36.

\bibitem[{Li et~al.(2023)Li, Holtzman, Fried, Liang, Eisner, Hashimoto, Zettlemoyer, and Lewis}]{li-etal-2023-contrastive}
Xiang~Lisa Li, Ari Holtzman, Daniel Fried, Percy Liang, Jason Eisner, Tatsunori Hashimoto, Luke Zettlemoyer, and Mike Lewis. 2023.
\newblock \href {https://doi.org/10.18653/v1/2023.acl-long.687} {Contrastive decoding: Open-ended text generation as optimization}.
\newblock In \emph{Proceedings of the 61st Annual Meeting of the Association for Computational Linguistics (Volume 1: Long Papers)}, pages 12286--12312, Toronto, Canada. Association for Computational Linguistics.

\bibitem[{Liu et~al.(2024)Liu, Qin, Ye, Mou, He, and Wang}]{liu2024adaptive}
Qingyi Liu, Jinghui Qin, Wenxuan Ye, Hao Mou, Yuxuan He, and Keze Wang. 2024.
\newblock Adaptive prompt routing for arbitrary text style transfer with pre-trained language models.
\newblock In \emph{Proceedings of the AAAI Conference on Artificial Intelligence}, volume~38, pages 18689--18697.

\bibitem[{Logacheva et~al.(2022)Logacheva, Dementieva, Ustyantsev, Moskovskiy, Dale, Krotova, Semenov, and Panchenko}]{logacheva-etal-2022-paradetox}
Varvara Logacheva, Daryna Dementieva, Sergey Ustyantsev, Daniil Moskovskiy, David Dale, Irina Krotova, Nikita Semenov, and Alexander Panchenko. 2022.
\newblock \href {https://doi.org/10.18653/v1/2022.acl-long.469} {{P}ara{D}etox: Detoxification with parallel data}.
\newblock In \emph{Proceedings of the 60th Annual Meeting of the Association for Computational Linguistics (Volume 1: Long Papers)}, pages 6804--6818, Dublin, Ireland. Association for Computational Linguistics.

\bibitem[{Luo et~al.(2023)Luo, Han, Mou, and Firdaus}]{luo-etal-2023-prompt}
Guoqing Luo, Yu~Han, Lili Mou, and Mauajama Firdaus. 2023.
\newblock \href {https://doi.org/10.18653/v1/2023.findings-emnlp.381} {Prompt-based editing for text style transfer}.
\newblock In \emph{Findings of the Association for Computational Linguistics: EMNLP 2023}, pages 5740--5750, Singapore. Association for Computational Linguistics.

\bibitem[{Madaan et~al.(2020)Madaan, Setlur, Parekh, Poczos, Neubig, Yang, Salakhutdinov, Black, and Prabhumoye}]{madaan-etal-2020-politeness}
Aman Madaan, Amrith Setlur, Tanmay Parekh, Barnabas Poczos, Graham Neubig, Yiming Yang, Ruslan Salakhutdinov, Alan~W Black, and Shrimai Prabhumoye. 2020.
\newblock \href {https://doi.org/10.18653/v1/2020.acl-main.169} {Politeness transfer: A tag and generate approach}.
\newblock In \emph{Proceedings of the 58th Annual Meeting of the Association for Computational Linguistics}, pages 1869--1881, Online. Association for Computational Linguistics.

\bibitem[{Mai et~al.(2023)Mai, Jiang, and Deng}]{mai2023prefix}
Huiyu Mai, Wenhao Jiang, and Zhihong Deng. 2023.
\newblock Prefix-tuning based unsupervised text style transfer.
\newblock \emph{arXiv preprint arXiv:2310.14599}.

\bibitem[{Men{\'e}ndez et~al.(1997)Men{\'e}ndez, Pardo, Pardo, and Pardo}]{menendez1997jensen}
ML~Men{\'e}ndez, JA~Pardo, L~Pardo, and MC~Pardo. 1997.
\newblock The jensen-shannon divergence.
\newblock \emph{Journal of the Franklin Institute}, 334(2):307--318.

\bibitem[{Meta(2024)}]{llama3}
Meta. 2024.
\newblock Introducing meta llama 3: The most capable openly available llm to date.
\newblock \url{https://ai.meta.com/blog/meta-llama-3/}.

\bibitem[{Minaee et~al.(2024)Minaee, Mikolov, Nikzad, Chenaghlu, Socher, Amatriain, and Gao}]{minaee2024large}
Shervin Minaee, Tomas Mikolov, Narjes Nikzad, Meysam Chenaghlu, Richard Socher, Xavier Amatriain, and Jianfeng Gao. 2024.
\newblock Large language models: A survey.
\newblock \emph{arXiv preprint arXiv:2402.06196}.

\bibitem[{Morcos and Barrett(2018)}]{morcos2018blog}
Ari Morcos and David Barrett. 2018.
\newblock Understanding deep learning through neuron deletion.
\newblock Google Deepmind Blog.
\newblock Https://deepmind.google/discover/blog/understanding-deep-learning-through-neuron-deletion.

\bibitem[{Mukherjee et~al.(2024{\natexlab{a}})Mukherjee, Bansal, Ojha, McCrae, and Du{\v{s}}ek}]{mukherjee2024text}
Sourabrata Mukherjee, Akanksha Bansal, Atul~Kr Ojha, John~P McCrae, and Ond{\v{r}}ej Du{\v{s}}ek. 2024{\natexlab{a}}.
\newblock Text detoxification as style transfer in english and hindi.
\newblock \emph{arXiv preprint arXiv:2402.07767}.

\bibitem[{Mukherjee et~al.(2024{\natexlab{b}})Mukherjee, Ojha, Bansal, Alok, McCrae, and Du{\v{s}}ek}]{mukherjee2024multilingual}
Sourabrata Mukherjee, Atul~Kr Ojha, Akanksha Bansal, Deepak Alok, John~P McCrae, and Ond{\v{r}}ej Du{\v{s}}ek. 2024{\natexlab{b}}.
\newblock Multilingual text style transfer: Datasets \& models for indian languages.
\newblock \emph{arXiv preprint arXiv:2405.20805}.

\bibitem[{Mukherjee et~al.(2024{\natexlab{c}})Mukherjee, Ojha, and Dušek}]{mukherjee2024are}
Sourabrata Mukherjee, Atul~Kr. Ojha, and Ondřej Dušek. 2024{\natexlab{c}}.
\newblock Are large language models actually good at text style transfer?
\newblock \emph{arXiv preprint arXiv:2406.05885}.

\bibitem[{Niu et~al.(2024)Niu, Liu, Zhu, and Penn}]{niu2024does}
Jingcheng Niu, Andrew Liu, Zining Zhu, and Gerald Penn. 2024.
\newblock What does the knowledge neuron thesis have to do with knowledge?
\newblock \emph{arXiv preprint arXiv:2405.02421}.

\bibitem[{Ostheimer et~al.(2023)Ostheimer, Nagda, Kloft, and Fellenz}]{ostheimer2023text}
Phil Ostheimer, Mayank Nagda, Marius Kloft, and Sophie Fellenz. 2023.
\newblock Text style transfer evaluation using large language models.
\newblock \emph{arXiv preprint arXiv:2308.13577}.

\bibitem[{Pan et~al.(2024)Pan, Lan, Li, and Qian}]{pan2024unsupervised}
Lei Pan, Yunshi Lan, Yang Li, and Weining Qian. 2024.
\newblock Unsupervised text style transfer via llms and attention masking with multi-way interactions.
\newblock \emph{arXiv preprint arXiv:2402.13647}.

\bibitem[{Radford et~al.(2019)Radford, Wu, Child, Luan, Amodei, Sutskever et~al.}]{radford2019language}
Alec Radford, Jeffrey Wu, Rewon Child, David Luan, Dario Amodei, Ilya Sutskever, et~al. 2019.
\newblock Language models are unsupervised multitask learners.
\newblock \emph{OpenAI blog}, 1(8):9.

\bibitem[{Rao and Tetreault(2018)}]{rao-tetreault-2018-dear}
Sudha Rao and Joel Tetreault. 2018.
\newblock \href {https://doi.org/10.18653/v1/N18-1012} {Dear sir or madam, may {I} introduce the {GYAFC} dataset: Corpus, benchmarks and metrics for formality style transfer}.
\newblock In \emph{Proceedings of the 2018 Conference of the North {A}merican Chapter of the Association for Computational Linguistics: Human Language Technologies, Volume 1 (Long Papers)}, pages 129--140, New Orleans, Louisiana. Association for Computational Linguistics.

\bibitem[{Schuster et~al.(2022)Schuster, Fisch, Gupta, Dehghani, Bahri, Tran, Tay, and Metzler}]{schuster2022confident}
Tal Schuster, Adam Fisch, Jai Gupta, Mostafa Dehghani, Dara Bahri, Vinh Tran, Yi~Tay, and Donald Metzler. 2022.
\newblock Confident adaptive language modeling.
\newblock \emph{Advances in Neural Information Processing Systems}, 35:17456--17472.

\bibitem[{Sellam et~al.(2020)Sellam, Das, and Parikh}]{sellam-etal-2020-bleurt}
Thibault Sellam, Dipanjan Das, and Ankur Parikh. 2020.
\newblock \href {https://doi.org/10.18653/v1/2020.acl-main.704} {{BLEURT}: Learning robust metrics for text generation}.
\newblock In \emph{Proceedings of the 58th Annual Meeting of the Association for Computational Linguistics}, pages 7881--7892, Online. Association for Computational Linguistics.

\bibitem[{Shazeer(2020)}]{shazeer2020glu}
Noam Shazeer. 2020.
\newblock Glu variants improve transformer.
\newblock \emph{arXiv preprint arXiv:2002.05202}.

\bibitem[{Shen et~al.(2017)Shen, Lei, Barzilay, and Jaakkola}]{shen2017style}
Tianxiao Shen, Tao Lei, Regina Barzilay, and Tommi Jaakkola. 2017.
\newblock Style transfer from non-parallel text by cross-alignment.
\newblock \emph{Advances in neural information processing systems}, 30.

\bibitem[{Tan et~al.(2024)Tan, Wu, and Monz}]{tan2024neuron}
Shaomu Tan, Di~Wu, and Christof Monz. 2024.
\newblock Neuron specialization: Leveraging intrinsic task modularity for multilingual machine translation.
\newblock \emph{arXiv preprint arXiv:2404.11201}.

\bibitem[{Tang et~al.(2024)Tang, Luo, Huang, Zhang, Wang, Zhao, Wei, and Wen}]{tang2024language}
Tianyi Tang, Wenyang Luo, Haoyang Huang, Dongdong Zhang, Xiaolei Wang, Xin Zhao, Furu Wei, and Ji-Rong Wen. 2024.
\newblock Language-specific neurons: The key to multilingual capabilities in large language models.
\newblock \emph{arXiv preprint arXiv:2402.16438}.

\bibitem[{Tigges et~al.(2023)Tigges, Hollinsworth, Geiger, and Nanda}]{tigges2023linear}
Curt Tigges, Oskar~John Hollinsworth, Atticus Geiger, and Neel Nanda. 2023.
\newblock Linear representations of sentiment in large language models.
\newblock \emph{arXiv preprint arXiv:2310.15154}.

\bibitem[{Touvron et~al.(2023)Touvron, Martin, Stone, Albert, Almahairi, Babaei, Bashlykov, Batra, Bhargava, Bhosale et~al.}]{touvron2023llama}
Hugo Touvron, Louis Martin, Kevin Stone, Peter Albert, Amjad Almahairi, Yasmine Babaei, Nikolay Bashlykov, Soumya Batra, Prajjwal Bhargava, Shruti Bhosale, et~al. 2023.
\newblock Llama 2: Open foundation and fine-tuned chat models.
\newblock \emph{arXiv preprint arXiv:2307.09288}.

\bibitem[{Vaswani et~al.(2017)Vaswani, Shazeer, Parmar, Uszkoreit, Jones, Gomez, Kaiser, and Polosukhin}]{vaswani2017attention}
Ashish Vaswani, Noam Shazeer, Niki Parmar, Jakob Uszkoreit, Llion Jones, Aidan~N Gomez, {\L}ukasz Kaiser, and Illia Polosukhin. 2017.
\newblock Attention is all you need.
\newblock \emph{Advances in neural information processing systems}, 30.

\bibitem[{Voigt et~al.(2018)Voigt, Jurgens, Prabhakaran, Jurafsky, and Tsvetkov}]{voigt-etal-2018-rtgender}
Rob Voigt, David Jurgens, Vinodkumar Prabhakaran, Dan Jurafsky, and Yulia Tsvetkov. 2018.
\newblock \href {https://aclanthology.org/L18-1445} {{R}t{G}ender: A corpus for studying differential responses to gender}.
\newblock In \emph{Proceedings of the Eleventh International Conference on Language Resources and Evaluation ({LREC} 2018)}, Miyazaki, Japan. European Language Resources Association (ELRA).

\bibitem[{Wang et~al.(2024)Wang, Wu, Rosa, Wang, and Shrivastava}]{wang2024multimodality}
Hanyu Wang, Pengxiang Wu, Kevin~Dela Rosa, Chen Wang, and Abhinav Shrivastava. 2024.
\newblock Multimodality-guided image style transfer using cross-modal gan inversion.
\newblock In \emph{Proceedings of the IEEE/CVF Winter Conference on Applications of Computer Vision}, pages 4976--4985.

\bibitem[{Wang et~al.(2022)Wang, Wen, Zhang, Hou, Liu, and Li}]{wang-etal-2022-finding-skill}
Xiaozhi Wang, Kaiyue Wen, Zhengyan Zhang, Lei Hou, Zhiyuan Liu, and Juanzi Li. 2022.
\newblock \href {https://doi.org/10.18653/v1/2022.emnlp-main.765} {Finding skill neurons in pre-trained transformer-based language models}.
\newblock In \emph{Proceedings of the 2022 Conference on Empirical Methods in Natural Language Processing}, pages 11132--11152, Abu Dhabi, United Arab Emirates. Association for Computational Linguistics.

\bibitem[{Xiao et~al.(2024)Xiao, Zhou, Ping, Cao, Li, Zhou, Li, and Bogdan}]{xiao2024exploring}
Xiongye Xiao, Chenyu Zhou, Heng Ping, Defu Cao, Yaxing Li, Yizhuo Zhou, Shixuan Li, and Paul Bogdan. 2024.
\newblock Exploring neuron interactions and emergence in llms: From the multifractal analysis perspective.
\newblock \emph{arXiv preprint arXiv:2402.09099}.

\bibitem[{Xu et~al.(2012)Xu, Ritter, Dolan, Grishman, and Cherry}]{xu-etal-2012-paraphrasing}
Wei Xu, Alan Ritter, Bill Dolan, Ralph Grishman, and Colin Cherry. 2012.
\newblock \href {https://aclanthology.org/C12-1177} {Paraphrasing for style}.
\newblock In \emph{Proceedings of {COLING} 2012}, pages 2899--2914, Mumbai, India. The COLING 2012 Organizing Committee.

\bibitem[{Yang et~al.(2024)Yang, Yu, Tian, Yan, Ma, and Zhang}]{yang2024evolutionary}
Shangshang Yang, Xiaoshan Yu, Ye~Tian, Xueming Yan, Haiping Ma, and Xingyi Zhang. 2024.
\newblock Evolutionary neural architecture search for transformer in knowledge tracing.
\newblock \emph{Advances in Neural Information Processing Systems}, 36.

\bibitem[{Zhang et~al.(2024)Zhang, Cai, Wu, Hou, Abdul-Mageed et~al.}]{zhang2024distilling}
Chiyu Zhang, Honglong Cai, Yuexin Wu, Le~Hou, Muhammad Abdul-Mageed, et~al. 2024.
\newblock Distilling text style transfer with self-explanation from llms.
\newblock \emph{arXiv preprint arXiv:2403.01106}.

\bibitem[{Zhao et~al.(2024)Zhao, Zhang, Chen, Kawaguchi, and Bing}]{zhao2024large}
Yiran Zhao, Wenxuan Zhang, Guizhen Chen, Kenji Kawaguchi, and Lidong Bing. 2024.
\newblock How do large language models handle multilingualism?
\newblock \emph{arXiv preprint arXiv:2402.18815}.

\end{thebibliography}

\clearpage
\appendix

\section{Datasets}
\label{app:dataset}
All style data used for neuron identification are obtained from publicly available datasets.
We applied the following preprocessing to the raw data:
(1) removing sentences longer than 120 characters;
(2) eliminating duplicate sentences; and
(3) removing sentences containing a large number of special symbols.
Table~\ref{tab:data_statistic} provides detailed statistics of the preprocessed corpus.

\section{Classifiers used in Each Benchmark}
\label{app:cls}
To evaluate the accuracy of style transfer, we use open-source classifiers on the six benchmarks we evaluated. The sources of these classifiers are detailed in Table~\ref{tab:cls_statistic}.

\section{JSD Distance between Layers}
\label{app:jsd}
To verify whether the style-specific layers selected in Section~\ref{sec:contras_decode} encode stylistic information, we calculate the Jensen-Shannon Divergence (JSD) distances between the final layer and all previous layers for the TST task of transfer from informal style text to formal style text.
The results, shown in Table~\ref{tab:jsd}, led to the following findings:
(1) For most of the early layers, from layer 0 to 26, the distances between the final layer and these layers remain almost constant or change very little, indicating that the information encoded in these layers is very similar.
However, for the last few layers, from layer $27$ to $31$, the JSD distance from the final layer is smaller compared to the earlier layers, but the distance between different layers increases.
This suggests that the last few layers are processing style-related information, consistent with the distribution characteristics of the style layers discussed in Section~\ref{sec:contras_decode}.
(2) Some words associated with the formal style (target-side style), highlighted in bold in the Table~\ref{tab:jsd}, show a larger distance difference in the last few layers.
This aligns with our expectation that words representing the target style are more likely to be activated in the style layer, increasing their probability of being selected as candidates for token generation in the style layer.

\section{Effectiveness of different model sizes}
\label{app:model_size}
To verify the effectiveness of our method on a larger model, we conduct experiments using the 70B version of LLaMA-3. The results, presented in Table~\ref{tab:res_70b}, indicate that our method is also effective on the larger model and consistent with the conclusions drawn from the 7B model (See Table~\ref{tab:main_res} in Section~\ref{sec:res} for more details).

%% Table style vs. our layer selection
\begin{table}[t]
\resizebox{\columnwidth}{!}{
\begin{tabular}{ll|cc}
\toprule
\multicolumn{2}{c}{\textbf{Style}}   & \textbf{Dola} & \textbf{Our}  \\
\midrule
\multirow{2}{*}{\textbf{Formality}}  & informal$\rightarrow$formal       & 78.14 & 80.80 \\
                            & formal$\rightarrow$informal       & 12.63 & 14.40 \\
\midrule
\multirow{2}{*}{\textbf{Toxicity}}   & toxic$\rightarrow$neutral         & 49.25 & 55.36 \\
                            & neutral$\rightarrow$toxic         & 25.41 & 31.98 \\
\midrule
\multirow{2}{*}{\textbf{Politics}}   & democratic$\rightarrow$republican & 36.26 & 37.81 \\
                            & republican$\rightarrow$democratic & 46.25 & 50.30 \\
\midrule
\multirow{2}{*}{\textbf{Politeness}} & impolite$\rightarrow$polite       & 76.58 & 80.63 \\
                            & polite$\rightarrow$impolite       & 20.57 & 23.27 \\
\midrule
\multirow{2}{*}{\textbf{Authorship}} & shakespeare$\rightarrow$modern    & 65.87 & 73.40 \\
                            & modern$\rightarrow$shakespeare    & 42.43 & 45.14 \\
\midrule
\multirow{2}{*}{\textbf{Sentiment}}  & positive$\rightarrow$negative     & 73.12 & 77.93 \\
                            & negative$\rightarrow$positive     & 50.28 & 54.73 \\
\bottomrule
\end{tabular}}
\caption{
\label{tab:dola_vs_our}
Comparison of different layer selection strategies between Dola and our approach.
}
\end{table}

\section{Style Layers vs. Dola Layers}
\label{app:style_layer}
In Section~\ref{sec:contras_decode}, our method selects the style layers, specifically the last few layers of the LLMs, to decoding from contrasting against the final layer.
In contrast, Dola selects the early layers to decode by contrasting the final layer.
To verify the superiority of our selected style layers, we conduct a comparison experiment, the results of which are shown in Table~\ref{tab:dola_vs_our}.
We can clearly observe the superiority of selecting the last few layers for contrastive decoding in the TST task.

\section{Different content preservation metrics}
\label{app:diff_content_prev}
In Table~\ref{tab:main_res}, we find that our method is not optimal in content preservation.
To verify whether this phenomenon occurs with other content preservation metric, we conduct a comparison experiment and present the results in Table~\ref{tab:mean_prev}.
We observe the same conclusion as in Table~\ref{tab:main_res}, namely, our method is inferior to the baseline method in terms of meaning preservation. For a detailed analysis, please refer to Section~\ref{sec:res}.

\section{Different decoding strategy}
\label{app:diff_decoding}
In Section~\ref{sec:contras_decode}, we present a decoding strategy for contrasting style layers.
To verify the advantages of this decoding strategy, we compare it with two additional decoding methods: nucleus sampling and contrastive search.
As shown in Table~\ref{tab:diff_decoding}, our decoding method outperforms the others.
This is primarily because contrastive search focuses on the isotropy of token representations during decoding, which means that the semantically similar words have less variation in the representation space and their probabilities should be increased.
However, this does not align with the goal of the TST task, which aims to expose more target-style words.
Source-style words and target-side style words are actually similar in representation.
For example, in the emotion task, ``like'' and ``hate'' are semantically different but similar in the embedding space because both represent an emotion, making it difficult to distinguish between these words using isotropy at the representation level.

In addition, nucleus sampling (NP) is a decoding method by setting a threshold $p$ and then restricting the sampling to the set of most probable tokens with cumulative probability less than $p$.
NP is not suited for TST because after deactivating the style neurons at the source side, the probability distribution of the words is changed.
The probability of all words in the target style becomes higher, resulting in candidate words predominantly being in the target style.
This can cause issues with fluency, as words in the target style are not always meant to be revealed in every context.

%% Table: datasets
\begin{table*}[htb]
\resizebox{\textwidth}{!}{
\begin{tabular}{lllccc}
\toprule
\multirow{2}{*}{\textbf{Benchmark}} & \multirow{2}{*}{\textbf{Dataset}} & \multirow{2}{*}{\textbf{Tasks}}                  & \multicolumn{3}{c}{\textbf{Size}} \\
\cmidrule(lr){4-6}
                           &                          &                                         & train   & vald   & test  \\
\midrule
Politeness                 & Politness~\cite{madaan-etal-2020-politeness}                & impolite $\leftrightarrow$ polite       & 100k    & 2000   & 2000  \\
Toxicity                   & ParaDetox~\cite{logacheva-etal-2022-paradetox}                & toxic $\leftrightarrow$ neutral         & 18k     & 2000   & 2000  \\
Formality                  & GYAFC~\cite{rao-tetreault-2018-dear}                    & informal $\leftrightarrow$ formal       & 52k     & 500    & 500   \\
Authorship                 & Shakespeare~\cite{xu-etal-2012-paraphrasing}              & shakespeare $\leftrightarrow$ modern    & 27k     & 500    & 500   \\
Politics                   & Political~\cite{voigt-etal-2018-rtgender}                & democratic $\leftrightarrow$ republican & 100k    & 1000   & 1000  \\
Sentiment                  & Yelp~\cite{shen2017style}                     & positive $\leftrightarrow$ negative     & 100k    & 1000   & 1000  \\
\bottomrule
\end{tabular}}
\caption{
\label{tab:data_statistic}
Data statistics on six benchmarks containing the size of train/valid/test set and transfer task we evaluated.
}
\end{table*}
%% Table -- classifier
\begin{table*}[htb]
\resizebox{\textwidth}{!}{
\begin{tabular}{l|l}
\toprule
\textbf{Benchmark}  & \textbf{Source}                                                                            \\
\midrule
Politeness & \url{https://huggingface.co/Genius1237/xlm-roberta-large-tydip}                         \\
Toxicity   & \url{https://huggingface.co/s-nlp/roberta\_toxicity\_classifier}                        \\
Formality  & \url{https://huggingface.co/s-nlp/xlmr\_formality\_classifier }                         \\
Authorship & \url{https://huggingface.co/notaphoenix/shakespeare\_classifier\_model}                 \\
Politics   & \url{https://huggingface.co/m-newhauser/distilbert-political-tweets}                   \\
Sentiment  & \url{https://huggingface.co/distilbert/distilbert-base-uncased-finetuned-sst-2-english} \\
\bottomrule
\end{tabular}
}
\caption{
\label{tab:cls_statistic}
Classifiers used to evaluate the accuracy of style transfer.
}
\end{table*}
%% Table --- JSD
\begin{table*}[!htp]
\resizebox{\textwidth}{!}{
\begin{tabular}{c|cccccccccccccccccccccc}
\toprule
& \multicolumn{22}{l}{
\makecell[l]{
\textbf{Instruction:} Please transfer the following informal style sentence into a formal style sentence and maintain the meaning of the sentence. \\
\textbf{Input:} the movie The In-Laws not exactly a holiday movie but funny and good! \\
\textbf{Output:} \textbf{The} movie \textbf{"}The In-Laws\textbf{"} \textbf{is} not exactly a holiday movie, but \textbf{it} \textbf{is} funny and good!
}
} \\
\midrule
   & \textbf{The}  & movie & \textbf{"}    & The  & In   & -L   & aws  & \textbf{"}    & \textbf{is}   & not  & exactly & a    & holiday & movie & ,    & but  & \textbf{it}   & \textbf{is}   & funny & and  & good & !    \\
\midrule
0  & 2.35 & 2.93  & 3.18 & 3.70 & 3.80 & 3.53 & 5.44 & 3.76 & 3.59 & 3.92 & 3.70    & 4.02 & 4.02    & 4.29  & 3.86 & 4.03 & 3.95 & 4.28 & 3.89  & 4.37 & 4.12 & 3.04 \\
1  & 2.35 & 2.93  & 3.18 & 3.68 & 3.79 & 3.52 & 5.42 & 3.74 & 3.58 & 3.91 & 3.70    & 4.02 & 4.01    & 4.28  & 3.86 & 4.02 & 3.93 & 4.27 & 3.87  & 4.36 & 4.11 & 3.04 \\
2  & 2.35 & 2.93  & 3.18 & 3.68 & 3.79 & 3.52 & 5.43 & 3.74 & 3.58 & 3.91 & 3.70    & 4.02 & 4.01    & 4.28  & 3.86 & 4.01 & 3.92 & 4.26 & 3.87  & 4.36 & 4.10 & 3.04 \\
3  & 2.35 & 2.93  & 3.18 & 3.68 & 3.79 & 3.52 & 5.43 & 3.74 & 3.58 & 3.91 & 3.68    & 4.02 & 4.01    & 4.27  & 3.87 & 4.01 & 3.92 & 4.26 & 3.87  & 4.36 & 4.10 & 3.04 \\
4  & 2.35 & 2.93  & 3.18 & 3.68 & 3.79 & 3.52 & 5.44 & 3.74 & 3.59 & 3.92 & 3.68    & 4.01 & 4.02    & 4.27  & 3.87 & 4.01 & 3.92 & 4.26 & 3.87  & 4.36 & 4.10 & 3.04 \\
5  & 2.35 & 2.93  & 3.18 & 3.67 & 3.79 & 3.52 & 5.44 & 3.75 & 3.59 & 3.90 & 3.68    & 4.02 & 4.02    & 4.28  & 3.87 & 3.99 & 3.91 & 4.26 & 3.87  & 4.37 & 4.10 & 3.05 \\
6  & 2.35 & 2.92  & 3.18 & 3.67 & 3.79 & 3.52 & 5.42 & 3.74 & 3.60 & 3.91 & 3.68    & 4.01 & 4.01    & 4.27  & 3.87 & 3.99 & 3.91 & 4.26 & 3.88  & 4.35 & 4.10 & 3.04 \\
7  & 2.35 & 2.92  & 3.18 & 3.67 & 3.78 & 3.50 & 5.43 & 3.75 & 3.58 & 3.90 & 3.67    & 3.99 & 4.01    & 4.28  & 3.87 & 3.99 & 3.90 & 4.24 & 3.87  & 4.33 & 4.10 & 3.04 \\
8  & 2.36 & 2.92  & 3.18 & 3.67 & 3.79 & 3.50 & 5.42 & 3.76 & 3.58 & 3.90 & 3.67    & 3.98 & 4.01    & 4.27  & 3.86 & 3.99 & 3.91 & 4.24 & 3.89  & 4.35 & 4.11 & 3.06 \\
9  & 2.36 & 2.92  & 3.18 & 3.67 & 3.77 & 3.50 & 5.42 & 3.75 & 3.58 & 3.90 & 3.68    & 3.99 & 4.01    & 4.26  & 3.86 & 4.01 & 3.91 & 4.24 & 3.87  & 4.33 & 4.11 & 3.05 \\
10 & 2.35 & 2.91  & 3.17 & 3.66 & 3.77 & 3.49 & 5.45 & 3.74 & 3.58 & 3.90 & 3.68    & 3.98 & 4.01    & 4.26  & 3.86 & 3.99 & 3.90 & 4.23 & 3.89  & 4.34 & 4.11 & 3.05 \\
11 & 2.35 & 2.91  & 3.16 & 3.65 & 3.76 & 3.48 & 5.44 & 3.75 & 3.58 & 3.89 & 3.68    & 3.97 & 4.00    & 4.27  & 3.86 & 3.99 & 3.89 & 4.23 & 3.89  & 4.35 & 4.12 & 3.05 \\
12 & 2.36 & 2.93  & 3.16 & 3.65 & 3.76 & 3.49 & 5.44 & 3.74 & 3.58 & 3.90 & 3.67    & 3.97 & 4.00    & 4.26  & 3.85 & 3.99 & 3.89 & 4.22 & 3.89  & 4.36 & 4.12 & 3.05 \\
13 & 2.35 & 2.93  & 3.17 & 3.66 & 3.76 & 3.50 & 5.44 & 3.73 & 3.58 & 3.89 & 3.68    & 3.97 & 4.02    & 4.27  & 3.89 & 3.98 & 3.89 & 4.23 & 3.91  & 4.35 & 4.12 & 3.06 \\
14 & 2.34 & 2.91  & 3.15 & 3.64 & 3.76 & 3.50 & 5.46 & 3.71 & 3.58 & 3.87 & 3.67    & 3.98 & 4.01    & 4.27  & 3.86 & 3.96 & 3.87 & 4.21 & 3.90  & 4.33 & 4.10 & 3.05 \\
15 & 2.34 & 2.90  & 3.14 & 3.62 & 3.78 & 3.50 & 5.44 & 3.71 & 3.55 & 3.86 & 3.66    & 3.97 & 4.01    & 4.26  & 3.84 & 3.93 & 3.87 & 4.20 & 3.91  & 4.31 & 4.11 & 3.04 \\
16 & 2.34 & 2.87  & 3.11 & 3.62 & 3.74 & 3.49 & 5.39 & 3.72 & 3.52 & 3.81 & 3.61    & 3.92 & 3.96    & 4.23  & 3.81 & 3.87 & 3.84 & 4.18 & 3.87  & 4.24 & 4.05 & 3.03 \\
17 & 2.32 & 2.87  & 3.08 & 3.61 & 3.71 & 3.46 & 5.39 & 3.71 & 3.53 & 3.79 & 3.60    & 3.89 & 3.93    & 4.21  & 3.79 & 3.74 & 3.81 & 4.17 & 3.85  & 4.22 & 4.02 & 2.99 \\
18 & 2.31 & 2.84  & 3.02 & 3.56 & 3.64 & 3.45 & 5.39 & 3.67 & 3.46 & 3.68 & 3.54    & 3.83 & 3.86    & 4.16  & 3.75 & 3.71 & 3.77 & 4.12 & 3.80  & 4.13 & 3.96 & 2.97 \\
19 & 2.30 & 2.80  & 3.00 & 3.53 & 3.61 & 3.42 & 5.38 & 3.62 & 3.41 & 3.62 & 3.47    & 3.78 & 3.84    & 4.13  & 3.71 & 3.67 & 3.68 & 4.07 & 3.77  & 4.09 & 3.92 & 2.93 \\
20 & 2.26 & 2.77  & 2.96 & 3.50 & 3.55 & 3.39 & 5.36 & 3.60 & 3.37 & 3.63 & 3.43    & 3.73 & 3.80    & 4.09  & 3.68 & 3.58 & 3.62 & 4.01 & 3.76  & 4.04 & 3.89 & 2.91 \\
21 & 2.23 & 2.74  & 2.92 & 3.46 & 3.50 & 3.39 & 5.33 & 3.60 & 3.30 & 3.50 & 3.39    & 3.65 & 3.78    & 4.04  & 3.62 & 3.44 & 3.58 & 3.95 & 3.72  & 3.93 & 3.84 & 2.87 \\
22 & 2.19 & 2.68  & 2.87 & 3.40 & 3.45 & 3.35 & 5.31 & 3.49 & 3.25 & 3.36 & 3.25    & 3.58 & 3.50    & 3.96  & 3.56 & 3.35 & 3.40 & 3.86 & 3.62  & 3.87 & 3.74 & 2.84 \\
23 & 2.14 & 2.57  & 2.80 & 3.33 & 3.35 & 3.33 & 5.27 & 3.44 & 3.15 & 3.28 & 3.11    & 3.47 & 3.34    & 3.88  & 3.49 & 3.25 & 3.28 & 3.73 & 3.54  & 3.77 & 3.61 & 2.81 \\
24 & 2.10 & 2.43  & 2.73 & 3.27 & 3.25 & 3.30 & 5.26 & 3.39 & 3.06 & 3.14 & 2.96    & 3.36 & 3.22    & 3.72  & 3.42 & 3.08 & 3.14 & 3.61 & 3.36  & 3.71 & 3.53 & 2.75 \\
25 & 2.07 & 2.37  & 2.60 & 3.22 & 3.16 & 3.25 & 5.24 & 3.33 & 2.96 & 3.02 & 2.77    & 3.22 & 2.71    & 3.65  & 3.34 & 3.03 & 3.00 & 3.54 & 3.20  & 3.60 & 3.38 & 2.70 \\
26 & 2.06 & 2.29  & 2.56 & 3.18 & 3.14 & 3.17 & 5.19 & 3.31 & 2.88 & 2.93 & 2.65    & 3.14 & 2.59    & 3.41  & 3.30 & 2.91 & 2.93 & 3.45 & 3.09  & 3.52 & 3.28 & 2.66 \\
\textbf{27} & 1.98 & 2.15  & 2.46 & 3.13 & 3.09 & 3.15 & 5.16 & 3.20 & 2.72 & 2.80 & 2.57    & 2.98 & 2.46    & 3.24  & 3.11 & 2.77 & 2.80 & 3.26 & 2.96  & 3.39 & 3.17 & 2.56 \\
\textbf{28} & 1.94 & 2.07  & 2.36 & 3.09 & 2.96 & 3.05 & 5.17 & 3.08 & 2.53 & 2.72 & 2.60    & 2.84 & 2.68    & 3.15  & 2.98 & 2.50 & 2.69 & 3.07 & 2.87  & 3.22 & 3.04 & 2.46 \\
\textbf{29} & 1.85 & 1.95  & 2.13 & 2.86 & 2.81 & 2.79 & 5.09 & 2.91 & 2.30 & 2.54 & 2.32    & 2.52 & 2.49    & 3.05  & 2.72 & 2.25 & 2.48 & 2.84 & 2.78  & 2.92 & 2.84 & 2.26 \\
\textbf{30} & 1.84 & 1.93  & 1.99 & 2.80 & 2.52 & 2.41 & 4.87 & 2.87 & 2.19 & 2.34 & 2.17    & 2.32 & 2.35    & 3.04  & 2.55 & 2.09 & 2.31 & 2.74 & 2.62  & 2.77 & 2.74 & 2.21 \\
\textbf{31} & 1.57 & 1.69  & 1.62 & 2.30 & 2.24 & 2.23 & 4.51 & 2.31 & 1.94 & 2.10 & 2.05    & 2.19 & 2.27    & 2.74  & 2.08 & 1.92 & 2.05 & 2.46 & 2.30  & 2.10 & 2.56 & 1.86 \\
\bottomrule
\end{tabular}
}
\caption{
\label{tab:jsd}
JSD (scaled by $10^5$) between the final layer and all previous layer in LLaMA-3. Each row represents the distance between all previous layers and the final layer, while each column corresponds to the token generated at each decoding step. Example taken from the TST task to transfer from informal style to formal style. The $0$-th layer is the embedding layer.
}
\end{table*}
%% Table -- model size
\begin{table*}[!th]
\resizebox{\textwidth}{!}{
\begin{tabular}{l|cccccccccccc}
\toprule
\multicolumn{13}{c}{\textbf{Style Transfer Accuracy}} \\
\midrule
 & \multicolumn{2}{c}{\textbf{Formality}} & \multicolumn{2}{c}{\textbf{Toxicity}} & \multicolumn{2}{c}{\textbf{Politics}} & \multicolumn{2}{c}{\textbf{Politeness}} & \multicolumn{2}{c}{\textbf{Authorship}} & \multicolumn{2}{c}{\textbf{Sentiment}} \\
\cmidrule(lr){2-3}\cmidrule(lr){4-5}\cmidrule(lr){6-7}\cmidrule(lr){8-9}\cmidrule(lr){10-11}\cmidrule(lr){12-13}
 & informal & formal & toxic  & neutral      & democratic    & republican   & impolite & polite & shakespeare    & modern & positive & negative     \\
\midrule
 & $\rightarrow$  & $\leftarrow$ & $\rightarrow$ & $\leftarrow$ & $\rightarrow$ & $\leftarrow$ & $\rightarrow$  & $\leftarrow$ & $\rightarrow$  & $\leftarrow$  & $\rightarrow$  & $\leftarrow$ \\
\midrule
LLaMA-3 & 81.45          & 11.73          & 48.57          & 30.57          & 35.97          & 49.43          & 81.21          & 15.05          & 64.55          & 44.67          & 77.06          & 53.08          \\
APE     & 75.22          & 13.39          & 49.43          & 28.62          & 42.83          & 46.40          & 78.32          & 19.64          & 56.07          & 45.22          & 79.20          & 48.64          \\
AVF     & 76.44          & 13.61          & 48.25          & 28.65          & 39.76          & 45.00          & 79.50          & 18.77          & 57.20          & 44.51          & 80.27          & 49.27          \\
PNMA    & 74.10          & 10.60          & 43.87          & 24.51          & 35.60          & 38.24          & 74.23          & 15.19          & 55.29          & 38.30          & 75.43          & 42.95          \\
Our     & \textbf{83.83} & \textbf{16.27} & \textbf{57.28} & \textbf{33.09} & \textbf{43.69} & \textbf{51.26} & \textbf{82.16} & \textbf{24.94} & \textbf{74.91} & \textbf{46.40} & \textbf{82.39} & \textbf{55.43} \\
\midrule
\midrule
\multicolumn{13}{c}{\textbf{Content Preservation}} \\
\midrule
 & \multicolumn{2}{c}{\textbf{Formality}} & \multicolumn{2}{c}{\textbf{Toxicity}} & \multicolumn{2}{c}{\textbf{Politics}} & \multicolumn{2}{c}{\textbf{Politeness}} & \multicolumn{2}{c}{\textbf{Authorship}} & \multicolumn{2}{c}{\textbf{Sentiment}} \\
\cmidrule(lr){2-3}\cmidrule(lr){4-5}\cmidrule(lr){6-7}\cmidrule(lr){8-9}\cmidrule(lr){10-11}\cmidrule(lr){12-13}
 & informal & formal & toxic  & neutral      & democratic    & republican   & impolite & polite & shakespeare    & modern & positive & negative     \\
\midrule
 & $\rightarrow$  & $\leftarrow$ & $\rightarrow$ & $\leftarrow$ & $\rightarrow$ & $\leftarrow$ & $\rightarrow$  & $\leftarrow$ & $\rightarrow$  & $\leftarrow$  & $\rightarrow$  & $\leftarrow$ \\
\midrule
LLaMA-3 & \textbf{86.29} & 76.54          & 73.85          & \textbf{84.59} & 83.01          & 77.49          & 76.03          & \textbf{90.99} & 79.89          & \textbf{64.10} & 76.63          & 74.89          \\
APE     & 77.26          & 85.28          & \textbf{78.18} & 83.33          & \textbf{89.48} & \textbf{83.52} & \textbf{77.28} & 88.09          & \textbf{81.72} & 59.37          & 76.62          & 74.06          \\
AVF     & 77.08          & \textbf{85.73} & 77.85          & 84.59          & 88.12          & 81.00          & 77.10          & 88.93          & 80.99          & 59.54          & \textbf{78.05} & 74.41          \\
PNMA    & 77.01          & 85.12          & 76.27          & 83.67          & 87.77          & 82.13          & 76.98          & 88.06          & 79.52          & 57.90          & 75.28          & 72.91          \\
Our     & 85.43          & 85.51          & 77.59          & 80.63          & 84.29          & 75.48          & 77.05          & 83.55          & 78.38          & 61.82          & 75.60          & \textbf{75.79} \\
\midrule
\midrule
\multicolumn{13}{c}{\textbf{Fluency}} \\
\midrule
 & \multicolumn{2}{c}{\textbf{Formality}} & \multicolumn{2}{c}{\textbf{Toxicity}} & \multicolumn{2}{c}{\textbf{Politics}} & \multicolumn{2}{c}{\textbf{Politeness}} & \multicolumn{2}{c}{\textbf{Authorship}} & \multicolumn{2}{c}{\textbf{Sentiment}} \\
\cmidrule(lr){2-3}\cmidrule(lr){4-5}\cmidrule(lr){6-7}\cmidrule(lr){8-9}\cmidrule(lr){10-11}\cmidrule(lr){12-13}
 & informal & formal & toxic  & neutral      & democratic    & republican   & impolite & polite & shakespeare    & modern & positive & negative     \\
\midrule
 & $\rightarrow$  & $\leftarrow$ & $\rightarrow$ & $\leftarrow$ & $\rightarrow$ & $\leftarrow$ & $\rightarrow$  & $\leftarrow$ & $\rightarrow$  & $\leftarrow$  & $\rightarrow$  & $\leftarrow$ \\
\midrule
LLaMA-3 & 87.93          & 86.58          & 112.55         & 190.29          & 86.41          & 63.86          & 101.40          & 90.29          & 196.47          & 135.93          & 172.68          & 122.16          \\
APE     & 93.56          & 89.34          & 128.33         & 187.91          & 87.70          & 65.33          & 105.00          & 91.75          & 246.71          & 131.55          & 146.58          & 123.43          \\
AVF     & 96.03          & 85.80          & 128.28         & 188.03          & 84.33          & 71.31          & 111.61          & 95.92          & 219.96          & 122.60          & 151.32          & 126.41          \\
PNMA    & 103.47         & 90.65          & 131.33         & 190.10          & 91.86          & 77.24          & 108.24          & 99.59          & 256.96          & 132.28          & 154.48          & 126.53          \\
Our     & \textbf{87.16} & \textbf{76.93} & \textbf{80.75} & \textbf{171.37} & \textbf{81.08} & \textbf{62.28} & \textbf{100.91} & \textbf{81.45} & \textbf{146.72} & \textbf{113.43} & \textbf{140.09} & \textbf{107.88} \\
\bottomrule
\end{tabular}
}
\caption{
\label{tab:res_70b}
\textbf{Main Results (70B model):} Style transfer accuracy (higher values are better; $\uparrow$), content preservation ($\uparrow$) and fluency ($\downarrow$) on $6$ datasets across $12$ transfer directions. Best results are highlighted in bold.
}

\end{table*}
%% table: content preservation
\begin{table*}[!t]
\resizebox{\textwidth}{!}{
\begin{tabular}{l|cccccccccccc}
\toprule
\multicolumn{13}{c}{\textbf{Content Preservation (Paraphrase Model)}} \\
\midrule
        & \multicolumn{2}{c}{\textbf{Formality}} & \multicolumn{2}{c}{\textbf{Toxicity}} & \multicolumn{2}{c}{\textbf{Politics}} & \multicolumn{2}{c}{\textbf{Politness}} & \multicolumn{2}{c}{\textbf{Authorship}} & \multicolumn{2}{c}{\textbf{Sentiment}} \\
\cmidrule(lr){2-3}\cmidrule(lr){4-5}\cmidrule(lr){6-7}\cmidrule(lr){8-9}\cmidrule(lr){10-11}\cmidrule(lr){12-13}
        & informal       & formal       & toxic         & neutral      & democratic    & republican   & impolite       & polite       & shakespeare    & modern        & positive       & negative     \\
\midrule
        & $\rightarrow$  & $\leftarrow$ & $\rightarrow$ & $\leftarrow$ & $\rightarrow$ & $\leftarrow$ & $\rightarrow$  & $\leftarrow$ & $\rightarrow$  & $\leftarrow$  & $\rightarrow$  & $\leftarrow$ \\
\midrule
LLaMA-3 & \textbf{85.95} & 74.71          & 73.54          & 82.71          & 82.48          & 75.77          & 75.32          & \textbf{89.14} & 78.75          & \textbf{62.28} & 76.17          & \textbf{74.47} \\
APE    & 76.72          & 85.06          & \textbf{76.72} & 83.00          & \textbf{87.99} & \textbf{82.21} & 76.80          & 87.89          & 80.07          & 57.61          & \textbf{76.52} & 73.53          \\
AVF    & 75.21          & 84.53          & 76.63          & \textbf{83.57} & 86.92          & 80.68          & \textbf{76.94} & 87.32          & \textbf{80.94} & 58.98          & 76.15          & 73.95          \\
PNMA    & 75.52          & 84.11          & 75.67          & 82.54          & 86.79          & 80.67          & 76.04          & 86.93          & 79.22          & 57.42          & 75.04          & 72.67          \\
Our     & 85.84          & \textbf{86.28} & 75.85          & 80.10          & 82.32          & 74.96          & 75.65          & 82.47          & 77.19          & 60.92          & 75.25          & 74.21          \\
\midrule
\multicolumn{13}{c}{\textbf{Content Preservation (LaBSE model)}} \\
\midrule
        & \multicolumn{2}{c}{\textbf{Formality}} & \multicolumn{2}{c}{\textbf{Toxicity}} & \multicolumn{2}{c}{\textbf{Politics}} & \multicolumn{2}{c}{\textbf{Politness}} & \multicolumn{2}{c}{\textbf{Authorship}} & \multicolumn{2}{c}{\textbf{Sentiment}} \\
\cmidrule(lr){2-3}\cmidrule(lr){4-5}\cmidrule(lr){6-7}\cmidrule(lr){8-9}\cmidrule(lr){10-11}\cmidrule(lr){12-13}
        & informal       & formal       & toxic         & neutral      & democratic    & republican   & impolite       & polite       & shakespeare    & modern        & positive       & negative     \\
\midrule
        & $\rightarrow$  & $\leftarrow$ & $\rightarrow$ & $\leftarrow$ & $\rightarrow$ & $\leftarrow$ & $\rightarrow$  & $\leftarrow$ & $\rightarrow$  & $\leftarrow$  & $\rightarrow$  & $\leftarrow$ \\
\midrule
LLaMA-3 & 0.75           & \textbf{0.90}  & 0.77           & 0.86           & 0.86           & 0.79           & 0.72           & \textbf{0.92}  & 0.86           & 0.70           & 0.76           & 0.76           \\
APE    & 0.75           & 0.88           & 0.78           & 0.87           & \textbf{0.90}  & 0.85           & 0.74           & 0.90           & 0.86           & 0.66           & 0.75           & 0.75           \\
AVF    & 0.74           & 0.88           & 0.78           & 0.87           & 0.89           & 0.83           & 0.75           & 0.90           & 0.87           & 0.67           & 0.75           & 0.76           \\
PNMA    & \textbf{0.79}  & 0.90           & \textbf{0.79}  & \textbf{0.89}  & 0.89           & \textbf{0.87}  & \textbf{0.86}  & 0.90           & \textbf{0.89}  & \textbf{0.74}  & \textbf{0.82}  & \textbf{0.81}  \\
Our     & 0.74           & 0.89           & 0.74           & 0.81           & 0.84           & 0.75           & 0.69           & 0.90           & 0.84           & 0.54           & 0.75           & 0.64           \\
\midrule
\multicolumn{13}{c}{\textbf{Content Preservation (BLEURT)}} \\
\midrule
        & \multicolumn{2}{c}{\textbf{Formality}} & \multicolumn{2}{c}{\textbf{Toxicity}} & \multicolumn{2}{c}{\textbf{Politics}} & \multicolumn{2}{c}{\textbf{Politness}} & \multicolumn{2}{c}{\textbf{Authorship}} & \multicolumn{2}{c}{\textbf{Sentiment}} \\
\cmidrule(lr){2-3}\cmidrule(lr){4-5}\cmidrule(lr){6-7}\cmidrule(lr){8-9}\cmidrule(lr){10-11}\cmidrule(lr){12-13}
        & informal       & formal       & toxic         & neutral      & democratic    & republican   & impolite       & polite       & shakespeare    & modern        & positive       & negative     \\
\midrule
        & $\rightarrow$  & $\leftarrow$ & $\rightarrow$ & $\leftarrow$ & $\rightarrow$ & $\leftarrow$ & $\rightarrow$  & $\leftarrow$ & $\rightarrow$  & $\leftarrow$  & $\rightarrow$  & $\leftarrow$ \\
\midrule
LLaMA-3 & \textbf{0.089} & \textbf{0.527} & 0.132          & 0.345          & 0.304          & 0.039          & \textbf{0.625} & 0.040          & 0.307          & 0.320          & 0.136          & 0.084          \\
APE    & 0.069          & 0.449          & 0.156          & \textbf{0.376} & \textbf{0.488} & \textbf{0.262} & 0.094          & \textbf{0.535} & 0.328          & \textbf{0.461} & 0.193          & 0.078          \\
AVF    & 0.043          & 0.440          & 0.157          & 0.376          & 0.424          & 0.191          & 0.122          & 0.522          & \textbf{0.344} & 0.426          & \textbf{0.207} & 0.095          \\
PNMA    & 0.002          & 0.433          & 0.139          & 0.360          & 0.399          & 0.181          & 0.074          & 0.513          & 0.334          & 0.417          & 0.197          & 0.085          \\
Our     & 0.073          & 0.478          & \textbf{0.157} & 0.329          & 0.460          & 0.232          & 0.557          & 0.473          & 0.324          & 0.386          & 0.199          & \textbf{0.133}       \\
\bottomrule
\end{tabular}
}
\caption{
\label{tab:mean_prev}
Different content preservation metrics: sentence embedding model trained from paraphrase datasets, sentence embedding model from multilingual representation model and BLEURT metrics.
}
\end{table*}
%%% table: different decoding
\begin{table*}[!th]
\resizebox{\textwidth}{!}{
\begin{tabular}{l|cccccccccccc}
\toprule
\multicolumn{13}{c}{\textbf{Style Transfer Accuracy}} \\
\midrule
        & \multicolumn{2}{c}{\textbf{Formality}} & \multicolumn{2}{c}{\textbf{Toxicity}} & \multicolumn{2}{c}{\textbf{Politics}} & \multicolumn{2}{c}{\textbf{Politeness}} & \multicolumn{2}{c}{\textbf{Authorship}} & \multicolumn{2}{c}{\textbf{Sentiment}} \\
\cmidrule(lr){2-3}\cmidrule(lr){4-5}\cmidrule(lr){6-7}\cmidrule(lr){8-9}\cmidrule(lr){10-11}\cmidrule(lr){12-13}
        & informal       & formal       & toxic         & neutral      & democratic    & republican   & impolite       & polite       & shakespeare    & modern        & positive       & negative     \\
\midrule
        & $\rightarrow$  & $\leftarrow$ & $\rightarrow$ & $\leftarrow$ & $\rightarrow$ & $\leftarrow$ & $\rightarrow$  & $\leftarrow$ & $\rightarrow$  & $\leftarrow$  & $\rightarrow$  & $\leftarrow$ \\
\midrule
NS ($p=0.95$) & 79.36 & 11.95 & 50.55 & 27.84 & 36.17 & 50.19 & 76.10 & 21.70 & 72.79 & 43.89 & 74.34 & 50.96  \\
CS    & 79.46 & 12.40 & 54.05 & 30.12 & 36.04 & 48.39 & 79.61 & 22.62 & 72.18 & 43.94 & 76.54 & 52.82 \\
Our    & \textbf{80.80} & \textbf{14.40} & \textbf{55.36} & \textbf{31.98} & \textbf{37.81} & \textbf{50.30} & \textbf{80.63} & \textbf{23.27} & \textbf{73.40} & \textbf{45.14} & \textbf{77.93} & \textbf{54.73}  \\
\bottomrule
\end{tabular}
}
\caption{
\label{tab:diff_decoding}
Comparison of three different decoding methods: nucleus sampling (NP; $p$=0.95), contrastive search (CS) and our decoding method.
}

\end{table*}

\end{document}